\documentclass[10pt,journal,compsoc]{IEEEtran}
\ifCLASSOPTIONcompsoc
 \usepackage[nocompress]{cite}
\else
 \usepackage{cite}
\fi
\usepackage{booktabs}
\ifCLASSINFOpdf
 \usepackage[pdftex]{graphicx}
 \graphicspath{}
\else
\fi
\usepackage{amsmath}
\usepackage{subfigure}
\usepackage{multicol}
\usepackage{graphicx}
\usepackage{dirtytalk}
\usepackage{hyperref}
\usepackage{multirow}
\usepackage[switch]{lineno}

\def\kformat{\textit{K}}  
\def\mformat{\textit{M}}  

\begin{document}
\title{FFD: Fast Feature Detector}
\author{Morteza~Ghahremani,~\IEEEmembership{Member,~IEEE}, ~Yonghuai~Liu,~\IEEEmembership{Senior Member,~IEEE}, and~Bernard~Tiddeman
\IEEEcompsocitemizethanks{\IEEEcompsocthanksitem M. Ghahremani and B. Tiddeman are with the Department of Computer Science, Aberystwyth University, UK, SY23 3DB.\protect\\
E-mail: \{mog9, bpt\}@aber.ac.uk
\IEEEcompsocthanksitem Y. Liu is with the Department of Computer Science, Edge Hill University, UK, L39 4QP.\protect\\
E-mail: yonghuai.liu@edgehill.ac.uk}
\thanks{Manuscript received x; revised x.

}}
\markboth{IEEE TRANSACTIONS ON IMAGE PROCESSING,\,~Vol.~x, No.~x, August~x}
{Shell \MakeLowercase{\textit{et al.}}: Bare Advanced Demo of IEEEtran.cls for IEEE Computer Society Journals}
\IEEEtitleabstractindextext{%
\begin{abstract}
Scale-invariance, good localization and robustness to noise and distortions are the main properties that a local feature detector should possess. 
Most existing local feature detectors find excessive unstable feature points that increase the number of keypoints to be matched and the computational time of the matching step.
In this paper, we show that robust and accurate keypoints exist in the specific scale-space domain. 
To this end, we first formulate the superimposition problem into a mathematical model and then derive a closed-form solution for multiscale analysis. The model is formulated via difference-of-Gaussian (DoG) kernels in the continuous scale-space domain, and it is proved that setting the scale-space pyramid's blurring ratio and smoothness to 2 and 0.627, respectively, facilitates the detection of reliable keypoints.
For the applicability of the proposed model to discrete images, we discretize it using the undecimated wavelet transform and the cubic spline function. 
Theoretically, the complexity of our method is less than 5\% of that of the popular baseline Scale Invariant Feature Transform (SIFT).
Extensive experimental results show the superiority of the proposed feature detector over the existing representative hand-crafted and learning-based techniques in accuracy and computational time. The code and supplementary materials can be found at~{\url{https://github.com/mogvision/FFD}}.
\end{abstract}
}
\maketitle

\IEEEdisplaynontitleabstractindextext
\IEEEpeerreviewmaketitle

\ifCLASSOPTIONcompsoc
\IEEEraisesectionheading{\section{Introduction}\label{sec:introduction}}
\else
\section{Introduction}
\label{sec:introduction}
\fi

\IEEEPARstart{F}{eature} detection is the process of extracting salient feature points from an image. 
The feature points could be blobs, corners or even edges~\cite{FusingRosten,Goodfeatures}.
Depending on the application, some operations are applied to the detected feature points. 
Feature detection finds numerous applications in the real world such as visual localization and 3D reconstruction. 
A good feature detector must provide reliable interest points/keypoints that are scale-invariant, highly distinguishable, robust to noise and distortions, valid with high repeatability rate, well localized, of easy implementation and computationally fast. 
Over the last three decades, a large number of image local feature detectors have been proposed, in which Scale Invariant Feature Transform (SIFT)~\cite{sift} is probably the most well-known technique and it actually opened a new era for image processing and computer vision. 
Since then, a considerable number of feature detectors have been proposed, where in most cases they followed and borrowed the concepts from SIFT like ~\cite{surf,asift,kaze,sifer,COSFIRE,brisk,palmprint,d2net}.

In the literature, feature detectors can be grouped into intensity-based, multiscale and learning-based categories. Intensity-based detectors are directly applied to the grey values of images. As expected, these detectors are usually fast. The Harris corner detector and its variants~\cite{harriscorner}, Features from Accelerated Segment Test (FAST)~\cite{fast1}, Maximally Stable Extremal Regions (MSER)~\cite{mser3}, Intensity-Based Regions (IBR)~\cite{ibm} and Smallest Uni-value Segment Assimilating Nucleus (SUSAN)~\cite{susan} are the most representative methods in this category.

The feature detectors of the second category use scale-space analysis. The input image is first transformed into a scale-space pyramid and then keypoints are detected. 
In the literature, such methods are often called multiscale feature detectors. 
Some representative multiscale feature detectors include SIFT, Speeded-Up Robust Feature (SURF)~\cite{surf}, Harris-Affine and Hessian-Affine~\cite{hessaff}, Affine SIFT (ASIFT)~\cite{asift}, a nonlinear scale-space method called KAZE\footnote[1]{KAZE means wind in Japanese and it stands for the nonlinear processes of the detector.}\cite{kaze}, Scale-Invariant Feature Detector with Error Resilience (SIFER)~\cite{sifer}, Combination Of Shifted FIlter REsponses (COSFIRE) \cite{COSFIRE} and multiscale Harris corner detector (HarrisZ)~\cite{harrisz}. 
Keypoints detected by the multiscale methods are usually of high accuracy, repeatability, robustness and scale-invariance. When compared to the intensity-based methods, they show better performance~\cite{Lindeberg3}, but usually require considerably more computational time.
In most applications, the feature detection step is followed by a feature description step and it is necessary to feed the descriptors with reliable keypoints, since reliable keypoints not only decrease the computational time of description but also increase the subsequent matching performance. 
Recently, several learnt feature detectors were developed~\cite{deepnet1,detnet,detdeep,lift,superpoint,d2net}.
In contrast with the methods in the former two categories, 
the methods in the third category do not extract and analyze particular features of the images for the identification of keypoints, but automatically learn and evaluate where they are and/or how they can be described. Even though such learning-based methods have the most potential, training data limits their applicability in practice. Other interesting feature detectors can be found in~\cite{censure,Brief,fastsift1,tal2016,MSFD,tip2}. 
Comprehensive surveys on local feature detectors are provided in~\cite{Lindeberg2,survey2017}.

Most of the conventional detectors cannot provide reliable keypoints and they usually fall in superimposed extrema while requiring considerable computational time. 
In this paper, we propose a novel multiscale feature detector for computer vision applications. 
Firstly, while the Difference-of-Gaussian (DoG) is often used to approximate the Laplacian of Gaussian (LoG), we analyze their relations in scale normalization and excitatory regions. 
The analysis reveals insights into the design of a suitable DoG kernel for feature detection in the continuous scale-space domain. 
This kernel ensures that the approximated LoG by the DoG is scale-normalized, the blurring ratio is optimized and the DoG will not produce superimposed extreme responses for the detection of keypoints in discrete images. 
The proposed kernel is then discretized for effective implementation using well-structured undecimated wavelets and the spline function to form our multiscale space domain. 
We search for reliable blobs laid at conjunctions via analysis of the hessian matrix and an anisotropic metric. 
The scale-space pyramid of the proposed method does not need either upsampling or downsampling operations and thus provides good localization for the detected keypoints. 
Theoretically, the computational time of the proposed feature detector is about 5\% that of SIFT while the keypoints detected by our detector are much more accurate and reliable than those of SIFT. 
Increasing reliability and reducing computational time considerably are the main characteristics of the proposed technique. 
For this reason, it is called fast feature detector, and for simplicity we abbreviate it as FFD. 

The rest of this paper is organized as follows. 
In the next section, we critically review existing feature detectors. 
The proposed fast feature detector is detailed in Section~\ref{sec::proposedmethod}.
Section~\ref{sec::experimental} reports and discusses the experimental results of FFD and the state-of-the-art keypoint detectors and, finally, conclusions and future work are drawn in Section~\ref{sec::conclusion}. %

\section{Related Work}
\label{sec::relatedwork}
In order to critically review SIFT\footnote[2]{Hereafter, for simplicity we denote the SIFT detector by SIFT, as the SIFT descriptor is not the study subject of this paper. 
The same notion is used for other methods.}, two issues should be considered: (\textit{i}) the framework of SIFT and (\textit{ii}) the methodology behind its implementation.
As discussed before, the framework of SIFT is well-established. 
It firstly transforms an input image into a suitable scale-space, which is scale-invariant (taking this feature into the design is important as we are interested in the scale-invariant keypoints in most applications), then in the scale-space, it detects interest blobs (candidate keypoints) and refines their locations in scale-space and, finally, rejects the unstable ones. 
As SIFT is in favour of blobs located at conjunctions, it computes the hessian matrix for each keypoint and selects the most reliable ones using a threshold on its eigenvalues. 
The majority of its computational time is assigned to the construction of its Gaussian scale-space pyramid. 
The blurring process of SIFT is slow, and aside from its high computational cost, it produces some unreliable keypoints due to its Gaussian smoothness.
Because of its scale-space, a considerable number of keypoints detected by SIFT are located over superimposed edges that lead to an increase in the running time of the descriptors and a decrease in matching performance subsequently.

Bay \textit{et al.}~\cite{surf} proposed a modified version of SIFT called SURF that approximates the Gaussian kernel by the integral image and Haar wavelets. While its computational time is significantly lower, approximated estimation of the Gaussian function seriously affects the localization and thus reliability of the detected keypoints. The same observation can be made in the results of BRISK~\cite{brisk}, which is a scale-invariant version of FAST.

To address the scale smoothing of SIFT, Alcantarilla \textit{et al.} \cite{kaze} proposed `KAZE'. 
This feature detector uses a nonlinear diffusion filter to form a nonlinear scale-space, and then detects the interest points. 
As it uses a nonlinear filter, it is robust to noise; moreover, as there is neither up-sampling nor down-sampling operation in its design, good localization is its another positive aspect. It, however, requires high computational time due to its nonlinear filter; to cope with this problem, its fast version under the name of `Accelerated KAZE (AKAZE)'~\cite{akaze} was proposed. 
The computational time is reduced but still high. Roughly speaking, AKAZE needs the same computational time as SIFT. 
Aside from the high complexity of KAZE and its accelerated version (this is because of the estimation of nonlinear filters), the detected keypoints often fall in superimposed extrema and their reliability against distortion is low.
Another improved version of KAZE is reported in~\cite{skaze}. 
A cosine modulated Gaussian filter was proposed in~\cite{sifer} to improve the performance of SIFT. According to the reported results, this method named SIFER enhanced the repeatability of the detected keypoints, but its computational time is considerably high and it seems to be unusable in practice. The same problem can be seen with techniques in~\cite{shearlets} and~\cite{wade}.

Recently, several deep learning-based feature detectors have been developed~\cite{deepnet1,detnet,detdeep,lift,superpoint,d2net}.
They train on patch-wise/full-sized images and often provide keypoints that are robust to distortion.
Even though the learning-based feature detectors have a certain degree of scale invariance because of pre-training with data augmentations~\cite{superpoint,hfnet}, they are not inherently invariant to scale changes and their matching tends to fail in cases with a significant change in scale.
In fact, data augmentation often captures well the variations in the real-world at the local level, but their effectiveness over large-scale datasets is usually difficult to predict.

Falling into superimposed extrema is the main problem of most existing feature detectors regardless of their categories.
The superimposition phenomenon is the interaction/interference between two or more adjacent edges in images whose kernel responses do not provide clear information about where these edges are. It happens in the cases that the parameters of the scale-space pyramids are not well defined.
In the following section, we will show that the reliable keypoints exist only in the specific scale-space and then reconstruct the proposed multiscale pyramid based on this. This study is the first attempt to solve the superimposed extrema problem for feature detection. 

\section{Proposed Fast Feature Detector (FFD)} \label{sec::proposedmethod}
Multiscale keypoint detectors generally contain two steps: scale-space pyramid construction and keypoint detection.
FFD is a multiscale feature detector for finding reliable blobs in images. 
We first need to design a suitable kernel for edge detection. In Section~\ref{sec::uwtlog}, we explore a new relationship between DoG and LoG kernels. This provides a solid foundation for designing our continuous scale-space. We prove that the parameters of the scale-space, i.e. smoothness and blurring ratio, could not be tuned arbitrarily. We explore the exact relation between the above-mentioned parameters by formulating the superimposition that occurred during edge detection. 
Section~\ref{sec::blussingrate} reveals that edges can be more reliably detected in the continuous scale-space if the blurring ratio and smoothness are set to 2 and 0.627, respectively. 
These are golden values for multiscale image processing. To the best of our knowledge, this is the first attempt to formalize and optimize the scale-space for detecting reliable edges in discrete images. We discretize our continuous scale-space using undecimated wavelet transform (UWT) and the spline function. We first review them in Section~\ref{sec::discretize} and then use them in the FFD multiscale architecture detailed in Section~\ref{sec::ffdarchitecture}. The last step of FFD, i.e. keypoints detection and refinement, is detailed in Section~\ref{sec::FeatureDetection}.

\subsection{Kernel Design in the Continuous Scale-Space}
\label{sec::uwtlog}
FFD is a blob-detector. 
The desired blob-detector kernel is the Laplacian of Gaussian (LoG). If a two-dimensional (2D) Gaussian function $G_{\sigma}(x,y)$ with width $\sigma$ is defined as
\begin{equation} \label{eq::appendixgaus}
G_{\sigma}(x,y)= \frac{1}{2\pi \sigma^2} e^{-\frac{x^2+y^2}{2\sigma^2}},
\end{equation}
then the scale-normalized LoG function is 
\begin{multline}\label{eq::LOG}
    \Bar{\bigtriangledown}^2 G_{\sigma}(x,y)=
    \sigma^2\bigtriangledown^2G_{\sigma}(x,y)\\
    =\sigma^2\big(\frac{\partial^2 G_{\sigma}(x,y)}{\partial x^2}+\frac{\partial^2 G_{\sigma}(x,y)}{\partial y^2}\big)
\\ = \frac{1}{\pi \sigma^2}\big(\frac{x^2+y^2}{ 2\sigma^2}-1\big)e^{-\frac{x^2+y^2}{2\sigma^2}},
\end{multline}
where $\bigtriangledown^2$ denotes the Laplacian operator in 2D space. In practice, LoG is not applicable to feature detection due to its high computational complexity and noise amplification since it contains the second-order derivative operations\footnote[1]{In practice, the input image is smoothed before applying the LoG kernel.}. 
Lindeberg~\cite{Lindeberg} and Lowe~\cite{sift} approximated an LoG function by a DoG one. They first replace time with scale in the heat diffusion
\begin{equation} \label{eq::heat}
 \frac{\partial G_{\sigma}(x,y)}{\partial \sigma} = \sigma \bigtriangledown^2G_{\sigma}(x,y).
\end{equation}
If the derivative of the Gaussian function is defined as
\begin{equation} \label{eq::heatII}
 \frac{\partial G_{\sigma}(x,y)}{\partial \sigma}=\lim_{\mu\sigma\to\sigma}{\frac{G_{\mu\sigma}(x,y)-G_{\sigma}(x,y)}{\mu\sigma- \sigma}},
\end{equation}
then Eq. (\ref{eq::heat}) can be approximated by
\begin{equation} \label{eq::heatfina2}
\Bar{\bigtriangledown}^2 G_{\sigma}(x,y)
\approx \frac{1}{\mu-1}\big(G_{\mu\sigma}(x,y)-G_\sigma(x,y)\big).
\end{equation}
Here `$G_{\mu\sigma}(x,y)-G_\sigma(x,y)$' denotes the DoG filter [hereafter it is denoted by $D_{\sigma, \mu}(x,y)$] and parameter $\mu$ is the ratio of two sigma values in the DoG function and is called blurring ratio. 
Equation~(\ref{eq::heatfina2}) states that the scale-normalized LoG function can be implemented by the DoG function. Compared to LoG, DoG is more robust to noise since it comprises two Gaussian filters that are inherently low-pass filters and thus can attenuate the side effects of noise. Moreover, the complexity of LoG is significantly reduced by DoG. 
However, this equation just provides an approximation for Eq. (\ref{eq::heat}) and could not describe the exact relation between the scale-normalized LoG and DoG functions.

To solve this problem, we define the exact relation between the scale-normalized LoG and DoG functions as follows
\begin{equation} \label{eq::heatfinal}
\Bar{\bigtriangledown}^2 G_{\sigma_L}(x,y)=\eta(\sigma_L,\sigma, \mu) D_{\sigma, \mu}(x,y),
\end{equation}
where $\sigma_L$ denotes the sigma value of the scale-normalized LoG kernel. 
In the above equation, $\eta(\sigma_L,\sigma, \mu)$ is a function that makes a balance between two sides of the aforementioned equation. 
Needless to say that $\eta(\sigma_L,\sigma, \mu)$ is approximated as `$\frac{1}{\mu-1}$' in Eq.~(\ref{eq::heatfina2}). In this study, we investigate its exact value.

We first assume that $\eta(\sigma_L,\sigma, \mu)$ is independent of $\sigma_L$ and $\sigma$, and then check whether this assumption is true or not. With this assumption and the linearity property of Eq. (\ref{eq::heatfinal}), the excitatory regions\footnote[2]{The excitatory region denoted by $\omega$ is the area enclosed by the two zero-crossing points in the second derivative-filters~\cite{edgedetection}. See Fig.~\ref{fig::dwd_a}.} of the DoG and the scale-normalized LoG kernels will be identical. 
The excitatory region of LoG, denoted by $\omega_L$, is obtained via setting Eq. (\ref{eq::LOG}) to zero:
\begin{equation} \label{eq::wL}
\omega_L=2\sqrt{x^2+y^2}=2\sqrt{2}\sigma_L.
\end{equation}
Likewise, one can formulate the excitatory region of DoG, $\omega_D$, as follows:
\begin{equation} \label{eq::wd}
\omega_D =4\mu\sigma\sqrt{\frac{\ln{(\mu)}}{\mu^2-1}}.
\end{equation}
Since the DoG and the scale-normalized LoG functions have the identical excitatory region, we can deduce
\begin{equation} \label{eq::sigmarelation}
\omega_L = \omega_D
\Rightarrow
\sigma_L=\mu\sigma \sqrt{\frac{2\ln{(\mu)}}{\mu^2-1}}.
\end{equation}
This equation reveals the relation between the locations of zero-crossing points in the DoG and the scale-normalized LoG functions. 
Now we investigate their amplitudes. 
The aim is to make the peaks of $\Bar{\bigtriangledown}^2 G_{\sigma}(x,y)$ and $D_{\sigma, \mu}(x,y)$ identical. 
The peak values of both the functions are situated at the centre, i.e. $x=0$ and $y=0$:
\begin{equation} \label{eq::etal}
\Bar{\bigtriangledown}^2 G_{\sigma_L}(0,0)=
-\frac{1}{\pi\sigma_L^2}=-\frac{\mu^2-1}{2\pi\mu^2\sigma^2\ln{(\mu})},
\end{equation}
and 
\begin{equation} \label{eq::etar}
D_{\sigma, \mu}(0,0)=
\frac{1}{2\pi\mu^2\sigma^2}-\frac{1}{2\pi\sigma^2}=-\frac{\mu^2-1}{2\pi\mu^2\sigma^2}.
\end{equation}
Inserting Eqs. (\ref{eq::etal}) and (\ref{eq::etar}) into Eq. (\ref{eq::heatfinal}) yields 
\begin{equation} \label{eq::etafinal}
\eta(\sigma_L,\sigma, \mu)=\frac{\Bar{\bigtriangledown}^2 G_{\sigma_L}(0,0)}{D_{\sigma, \mu}(0,0)}=\frac{1}{\ln{(\mu)}}.
\end{equation}
This equation clearly shows that $\eta(\sigma_L,\sigma, \mu)$ is independent of $\sigma_L$ and $\sigma$, as claimed earlier, and it just depends on the blurring ratio. Since the blurring ratio is always fixed in the scale-space pyramid, we may conclude that a DoG function with a sigma value of $\sigma$ and blurring ratio of $\mu$ is always scale-normalized at $\mu\sigma \sqrt{\frac{2\ln{(\mu)}}{\mu^2-1}}$ [see Eqs. (\ref{eq::heatfinal}) and (\ref{eq::sigmarelation})]. 
This is an important conclusion that the DoG function is scale-normalized under any conditions. We summarize the exact relation between the normalized LoG and DoG functions in the following:
\begin{equation} \label{eq::logdog}
\Bar{\bigtriangledown}^2 G_{\sigma_L}(x,y)=\frac{1}{\ln{(\mu)}} D_{\sigma, \mu}(x,y),
\end{equation}
where
\begin{equation} \label{eq::rd}
\sigma_L=\mu\sigma \sqrt{\frac{2\ln{(\mu)}}{\mu^2-1}}.
\end{equation}

If we seek the behaviour of the model defined in Eq. (\ref{eq::logdog}) for $\mu$ around 1, firstly we need to expand `$\ln{(\mu)}$' around `$\mu=1$' via the Taylor series
\begin{equation}\label{eq::lnmu}
\ln{(\mu)}= 
(\mu-1)-\frac{1}{2}(\mu-1)^2+...+\frac{(-1)^{N-1}}{N}(\mu-1)^N.
\end{equation}
Since $\mu$ approaches 1, it is possible to approximate the Taylor series of `$\ln{(\mu)}$' by its first term and, not surprisingly, it then yields `$\eta(\mu)\approx\frac{1}{\mu-1}$', exactly as stated earlier in Eq. (\ref{eq::heatfina2}). 
Unlike Eq. (\ref{eq::heatfina2}) that enforces the blurring ratio to be near 1, our model shows that this parameter can actually be chosen freely in the interval of $(1,\infty)$. Now, a question arises what suitable values for $\mu$ and $\sigma$ are. In the following subsection, we determine them using the superimposition concept and Eq. (\ref{eq::rd}).

\subsection{Determination of Blurring Ratio $\mu$ and Smoothness $\sigma$}\label{sec::blussingrate}
\begin{figure}
\unitlength .5cm
    \centering
    \subfigure[DoG kernels for different values of $\mu$.]{
    \includegraphics[width=\linewidth]{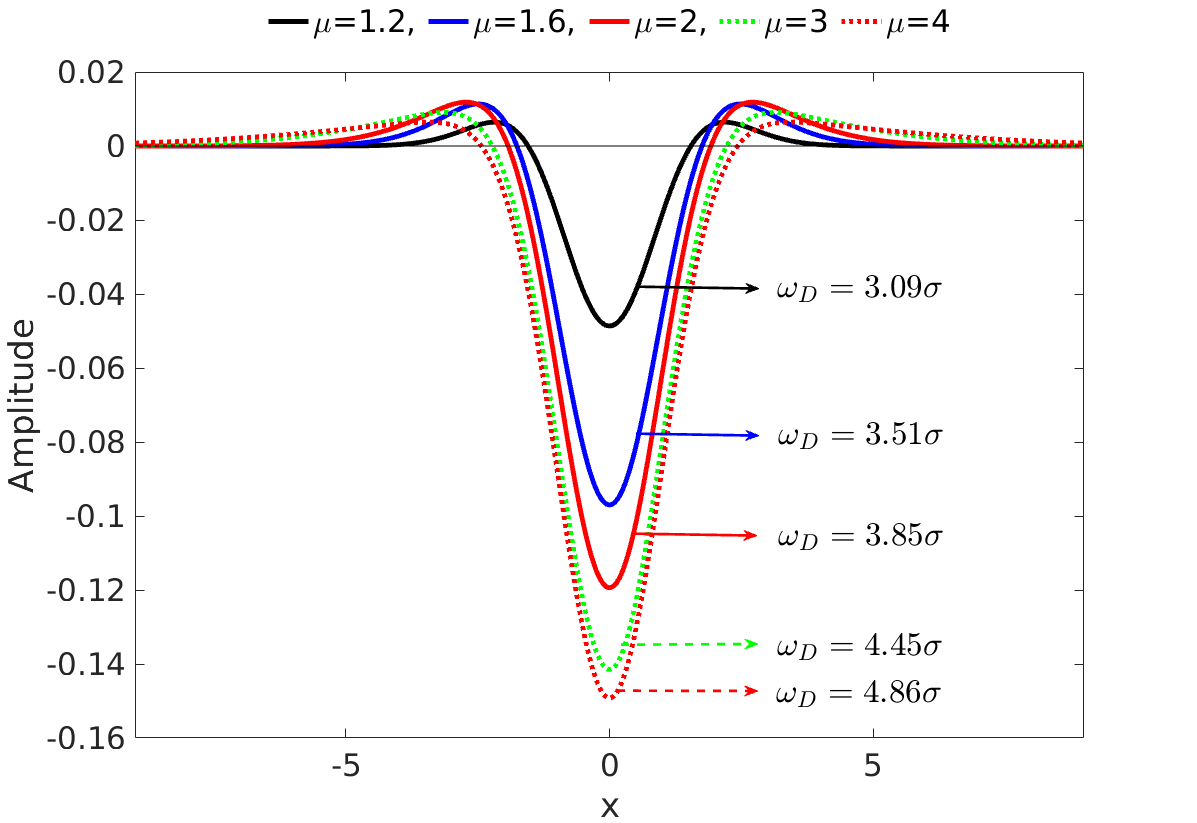}
    \label{fig::dogn}}
    \subfigure[Zero-crossing error for different values of $\mu$ and $\lambda$.]{
    \includegraphics[width=\linewidth] {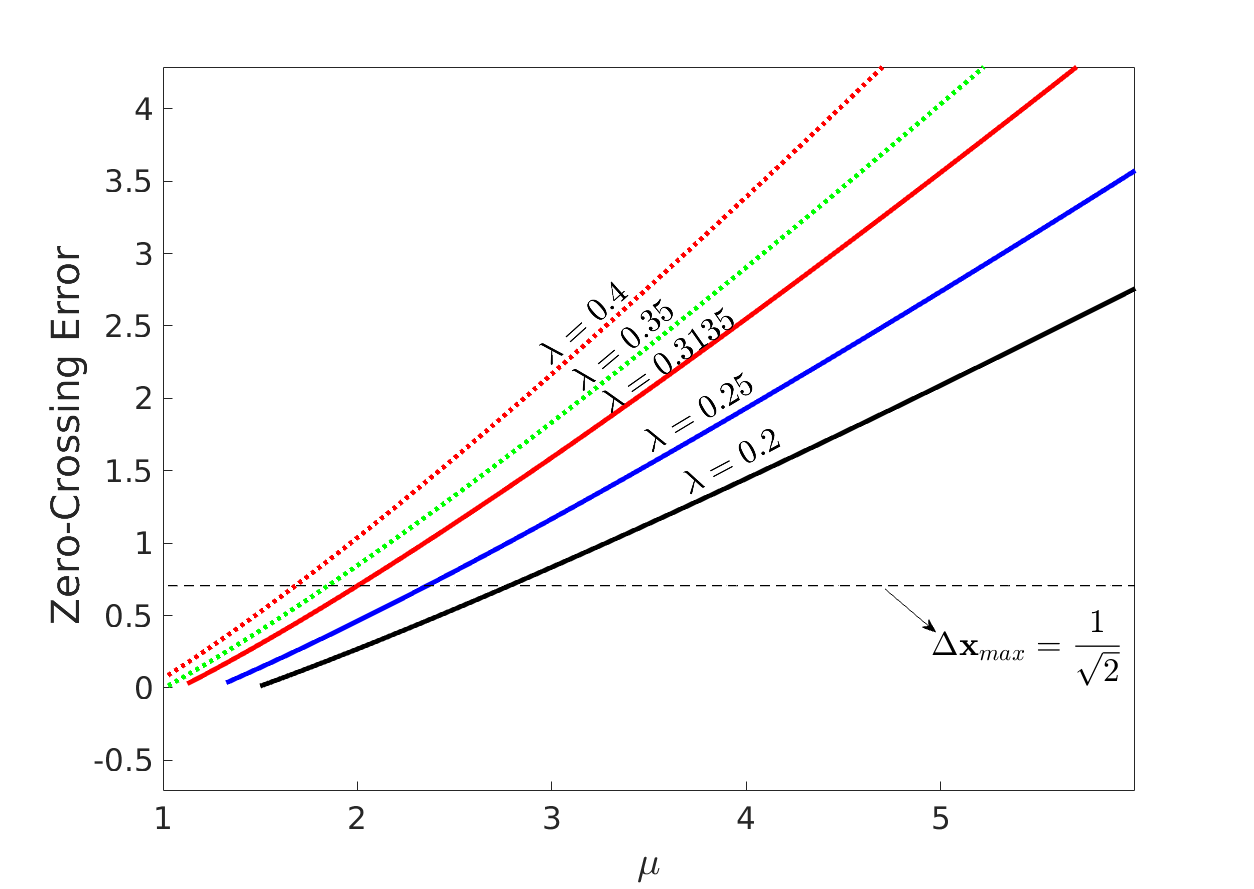}\label{fig::appendixBim}}
    \caption{(a) Different scale-ratio values $\mu$ yield different DoG functions, and subsequently, different excitatory regions. (b) Zero-crossing error for different values of $\mu$ and $\lambda$ according to Eq. (\ref{eq::Musigma}).}
\end{figure}

Parameter $\mu$ in Eq. (\ref{eq::logdog}) determines the scale-ratio, controlling the ratio of two sigma values in and thus the blurring speed of the DoG function. Fig. \ref{fig::dogn} depicts the DoG function for different values of $\mu$. 
It can be seen that the excitatory region is increased by raising the blurring ratio [recall Eq. (\ref{eq::rd})]. 
Determining the optimal DoG kernel has been a challenging task in image processing as the uncertainty theorem dictates in the space and the frequency domains. 
Marr and Hildreth~\cite{Marr} claimed that the suitable value of $\mu$ is 1.6. 
The authors in~\cite{Tinku} mentioned that the possible values for $\mu$ could be in the range of (1, 2]. 
Lowe~\cite{sift} considered `$\mu=2^{1/3}\approx1.26$' as the best value. However, all the aforementioned values of $\mu$ are obtained under some numerical experiments and there is no formal proof for their optimal determinations. 
In this study, we introduce a novel framework for thoroughly analyzing features in images. We will show that the blurring ratio plays a key role in the reliability of the detected keypoints.

\begin{figure}
\label{fig::dwd}
\unitlength .5cm
    \centering
    \subfigure[Input edges and a DoG kernel]{
    \includegraphics[width=0.55\linewidth]{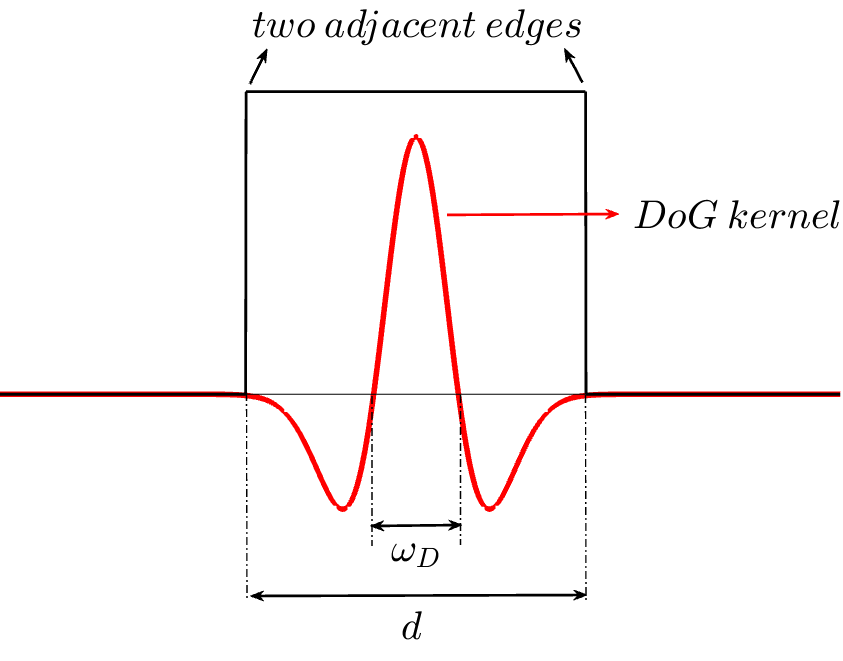}\label{fig::dwd_a}}
    \centering
    \subfigure[$d>3\omega_D$]{
    \includegraphics[width=0.5\linewidth]{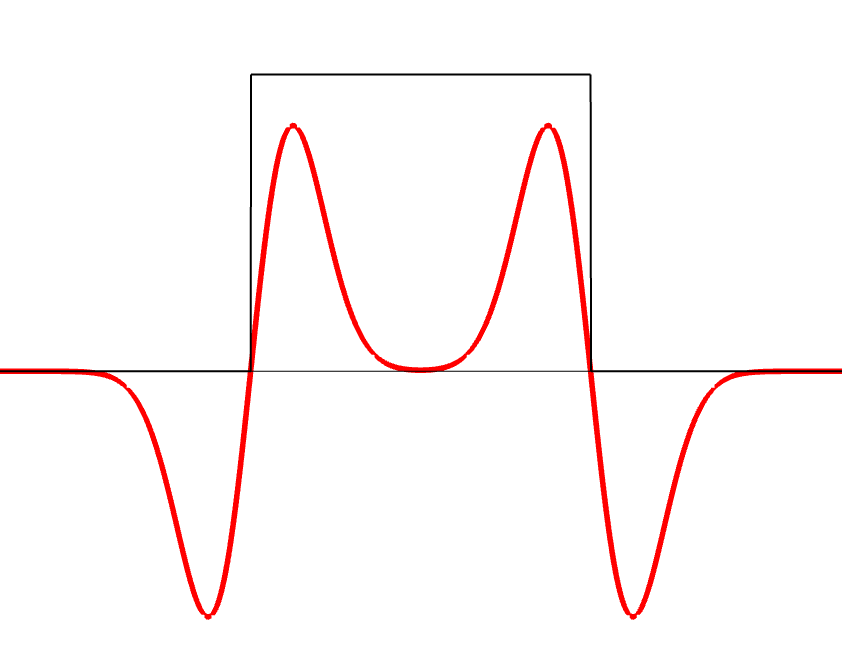}\label{fig::dwd_b}}
    \centering
    \subfigure[$\omega_D \leq d\leq3 \omega_D$]{
    \includegraphics[width=0.65\linewidth]{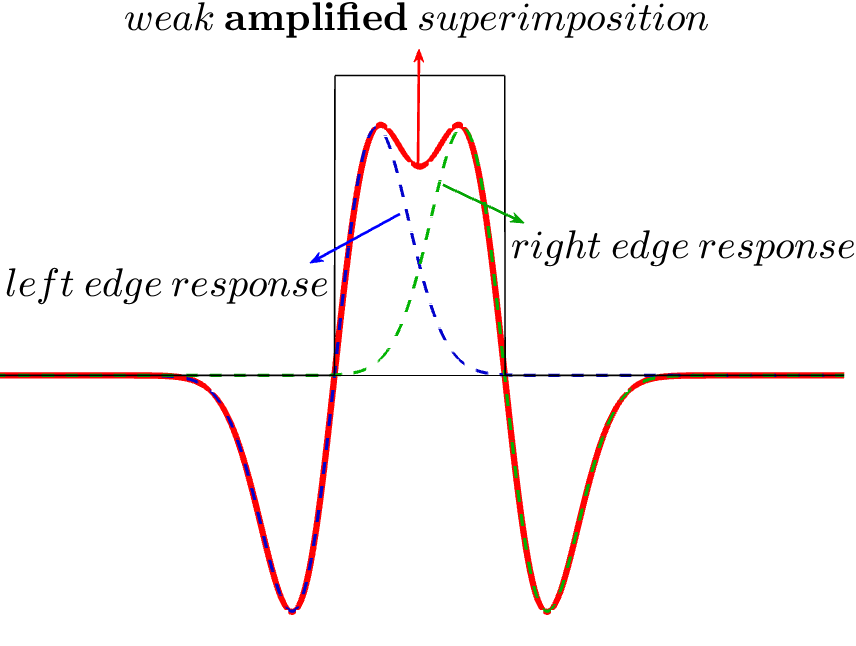} \label{fig::dwd_c}}
    \centering
    \subfigure[$d<\omega_D$]{
    \includegraphics[width=0.65\linewidth]{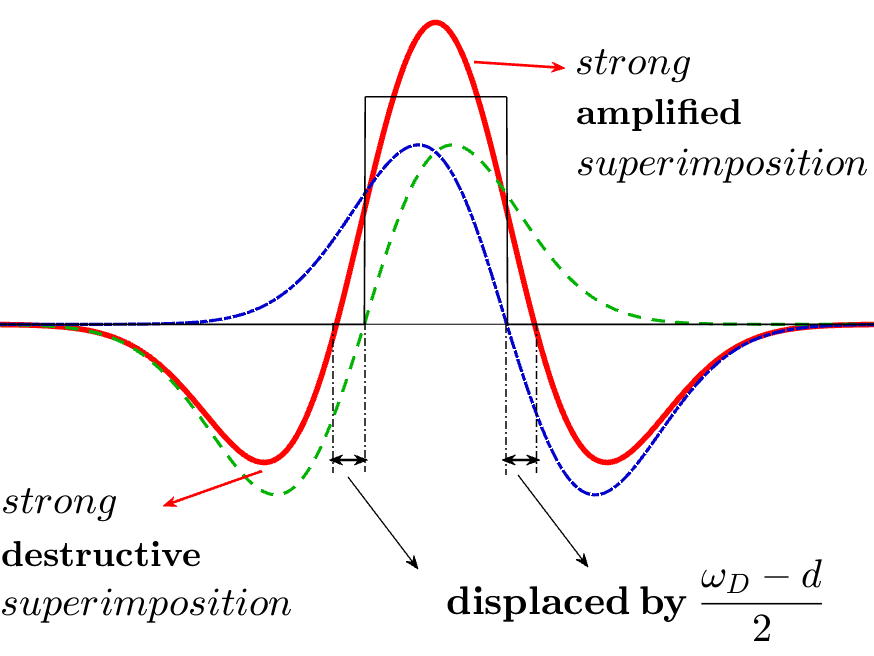}\label{fig::dwd_d}}
    \caption{DoG kernel responses are influenced by the relation between the kernel's excitatory region $\omega_D$ and the distance $d$ between two adjacent edges.}
\end{figure}

For simplicity, we discuss the DoG kernel in one dimension and the same results can be extended to two dimensions. 
When a kernel is applied to a signal, close edges may affect each other's responses. Depending on their distribution, their interaction could be destructive and amplified. Because of this interaction, superimposed edges do not show their real kernel responses so they are not reliable. 
Furthermore, superimposed edges may be displaced. 
In the image processing literature, this phenomenon is called superimposition~\cite{edgedetection}. A considerable number of detected blobs in the scale-space pyramid are caused by the superimposition.
Here, we formulate the superimposition problem based on $\mu$ and $\sigma$ and then derive a closed-form solution to their principled and optimal determination. 
The closeness of two adjacent edges is determined by the excitatory region $\omega_D$ of the DoG kernel. The support region is generally taken as $3\omega_D$, where 99.7\% of the area under a 1D Gaussian lies in `$3\omega_D$'~\cite{edgedetection}. If `$d$' denotes the distance between two adjacent edges [Fig.~\ref{fig::dwd_a}], then there are three possibilities:
\begin{itemize}
    \item $d>3\omega_D$: 
     The mutual influence of large-gap edges on the response to the DoG kernel is relatively weak and this can be ignored as depicted in Fig.~\ref{fig::dwd_b};
    
    \item $\omega_D\leq d\leq3\omega_D$: The mutual influence of medium-gap edges on the response to the DoG kernel is considerable [Fig.~\ref{fig::dwd_c}].
    
    \item $d<\omega_D$: The mutual influence of nearby edges on the response to the DoG kernel is so strong that it cannot determine their exact locations [Fig.~\ref{fig::dwd_d}].
\end{itemize}

If the width of interest region is smaller than the excitatory region of the applied kernel, i.e. $d<\omega_D$, the edges of the given region are displaced and their zero-crossing error $\Delta \mathbf{x}$ can be computed as follows:
\begin{equation}
\label{eq::appendixB1}
\Delta \mathbf{x}=\sqrt{\delta x^2+\delta y^2}=\frac{\omega_D-d}{2},
\end{equation}
where $\delta x$ and $\delta y$ are the deviation of the estimated zero-crossing point from its real value along $x$ and $y$ axes, respectively. 
The zero-crossing error increases when the difference between $\omega_D$ and $d$ increases and in the worst case, two adjacent edges have the minimum distance from each other. If this distance is denoted by $d_{min}$, then
\begin{equation}
\label{eq::appendixB2}
\Delta \mathbf{x}_{max}=\sqrt{\delta x_{max}^2+\delta y_{max}^2}=\frac{\omega_D-d_{min}}{2},
\end{equation}
where $\Delta \mathbf{x}_{max}$ is the maximum tolerable value of the zero-crossing error. 
The minimum distance between two adjacent edges is 1 pixel, i.e. `$d_{min}=1$'. 
On the other hand, the maximum deviation of an extremum along $x$ and $y$ axes should be less than $\frac{1}{2}$ pixel, i.e.  `$\delta x_{max}=\frac{1}{2}$' and `$\delta y_{max}=\frac{1}{2}$'. 
This is because deviations less than $\frac{1}{2}$ pixel could be refined (this will be discussed later in Eqs. (\ref{eq::taylor1}) and (\ref{eq::taylor2})), otherwise there is a shift in the location of the given pixel and we should check whether it is an extremum in the new location.
Considering these yields the following constraint on the excitatory region of the DoG kernel:
\begin{equation}
\label{eq::appendixBef}
\Delta \mathbf{x}_{max}\leq\frac{1}{\sqrt 2}
\Rightarrow
\frac{\omega_D-1}{2}\leq\frac{1}{\sqrt 2}.
\end{equation}
If we assume `$\sigma=\lambda\mu$' in Eq. (\ref{eq::rd}) where $\lambda$ is a positive constant, then inserting 
it into Eq. (\ref{eq::appendixBef}) gives the following constraint on $\lambda$ and $\mu$:
\begin{equation}
\label{eq::Musigma}
\frac{4\lambda\mu^2\sqrt{\frac{\ln{(\mu)}}{\mu^2-1}}-1}{2}\leq\frac{1}{\sqrt 2}.
\end{equation}
This constraint states that both parameters $\mu$ and $\sigma$ determine the zero-crossing error. 
In Fig. \ref{fig::appendixBim}, we depict the potential values of $\mu$ for different $\lambda$. For a fixed $\lambda$, increasing $\mu$ raises the zero-crossing error or, equivalently, the precision $\Delta \mathbf{x}$ in the space domain becomes coarse while the precision in the frequency domain (denoted by $\Delta{s}$) is enhanced according to the uncertainty theorem `$\Delta \mathbf{x} \Delta{s} \geq \frac{\pi}{4}$'~\cite{Marr}. 
As there is a trade-off between $\Delta \mathbf{x}$ and $\Delta{s}$, we need to select $\mu$ that satisfies Eq. (\ref{eq::Musigma}) and at the same time yields fine precision in the frequency domain. 
In our experiment, the DoG kernel for $\mu$ in the range of $(1, 3]$ has good similarity with its corresponding LoG kernel in the space domain as shown in Fig. \ref{fig::dogn}. 
On the other hand, if the bandwidth of the DoG kernel is analyzed in the frequency domain, one can compute that the half-power (-3dB) bandwidth for `$\mu=2$' is about $75.3\%$ of that at `$\mu=1+10^{-10}$'\footnote{According to Eq. (\ref{eq::heatfina2}), $\sigma_L$ is close to $\sigma$ when $\mu$ approaches 1. 
Here, we consider `$\mu=1+10^{-10}$' as the closest value to 1.}. 
Hence, $\mu$ in the range of $(1, 2]$ may provide reasonable bandwidth and we select $\mu=2$ for its good precision both in the frequency and space domains. 
By setting $\mu$ to 2, parameter $\lambda$ equals 0.3135 according to Eq. (\ref{eq::Musigma}) and this renders $0.627$ for $\sigma$ [see Fig. \ref{fig::appendixBim}]. 
In summary, $\sigma$ and $\mu$ must be carefully determined so that the responses of the DoG kernels can facilitate the separation of the nearby edges. Most conventional feature detectors have overlooked such considerations. Our analysis shows that setting $\sigma$ and $\mu$ as $0.627$ and 2 respectively guarantees no superimposed blobs. We thus construct our proposed multiscale space pyramid based on these golden values.

\subsection{Discrete Pyramid Design by the Golden Blurring Ratio and Gaussian Kernel}
\label{sec::discretize} 

Several studies~\cite{natural,Lindeberg} reveal that natural images have specific properties that exist over a certain range of scales, and the possible scale-space kernel is the Gaussian function. 
In practical image processing, a powerful transform is the one that can provide expressive representation for the structural information of an image.
The structural information is mainly edges and textures. 
In order to design a scale-space pyramid applicable to the discrete nature of the image using the DoG kernel with parameters `$\sigma=0.627$' and `$\mu=2$', we need to discretize the kernel. 
Hence, we seek a transform that can provide multiscale object representation with `$\mu=2$' and its kernel is similar to the Gaussian function in the discrete domain.
We are also interested in other aspects of a good transform including translation-invariance, good localization and robustness to noise and distortions. 
It is clear that a robust multiscale algorithm can provide stable representation for the structural information.

Taking all the above-mentioned factors into consideration motivates us to select UWT and the spline function at the heart of our feature detector. 
The scale-ratio of UWT is a constant of almost 2 that makes it more suitable for our design. 
Moreover, UWT is an undecimated transform and it is shown that redundant transforms are robust against noise and distortions~\cite{noiseredun}.
The spline kernel, on the other hand, is the approximation of the Gaussian function and is suitable for effective analysis of natural images. 
It is worth noting that there is a vast literature on different kernels. 
Haar, Daubechies, Biorthogonal, Coiflet, Symlet, Morlet, Mexican hat and different $B_N$-splines are probably the most applicable kernels in image processing. 
Likewise, there are a large number of studies on image/signal transforms. Wavelet and its numerous decimated and undecimated variants, platelet, ridgelet, curvelet, contourlets, bandlet, shearlet and ripplet are the most representative transforms. 
Surveying all of them is out of the scope of this paper and the reader can refer to~\cite{starckbook,mallat} and references therein for more information.

Here, we briefly review the UWT with a cubic spline finite impulse response (FIR) $B_3$ filter bank. 
The undecimated wavelet transform, which is also known as a stationary wavelet transform, is introduced by~\cite{UWT,starck2007}. 
This transform maps an image into different scale levels and then subtracts any two sequential scale/coarse images to yield the fine ones. 
If the kernel of UWT is the Gaussian function, then the fine scale of UWT is of the DoG functions. 
Instead of applying the downsampling operator to the input images, UWT upscales the kernel by a factor of $2^{j-1}$, where $j$ denotes the $j$th decomposition level of the image. 
The upsampling step is done via inserting zeros between the elements of the mother kernel and for this reason, this transform is also known as ``algorithme \`{a} trous''~\cite{atrous}. 
The UWT has a redundant framework of $J$ for $J$ decomposition levels that makes it robust to ringing artefacts.
At each scale level $j$, it extracts a coarse image $\mathbf{C}_j$ from its previous scale level `$j-1$'~\cite{starckbook}:
\begin{equation} \label{eq1}
\mathbf{C}_{j} = \mathbf{C}_{j-1}*h^{(j)}, \quad j=1,...,J;
\end{equation}
where $h^{(j)}$ denotes the kernel at scale level $j$ and $C_0$ is the input image. 
We construct our scale-space via the above equation and call it `coarse scale-space pyramid'. 
Subtraction of any two successive layers in the coarse scale-space pyramid yields the fine one $\mathbf{D}_j$ as
\begin{equation} \label{eq2}
\mathbf{D}_{j} = \mathbf{C}_{j-1} - \mathbf{C}_{j}, \quad j=1,...,J.
\end{equation}
Similarly, the `fine scale-space pyramid' includes all the fine images obtained via the above operation. 

\begin{figure}
\begin{minipage}[h]{1.0\linewidth}
\unitlength .5cm
    \centering
    \subfigure[]{
    \includegraphics[width=0.48\linewidth]{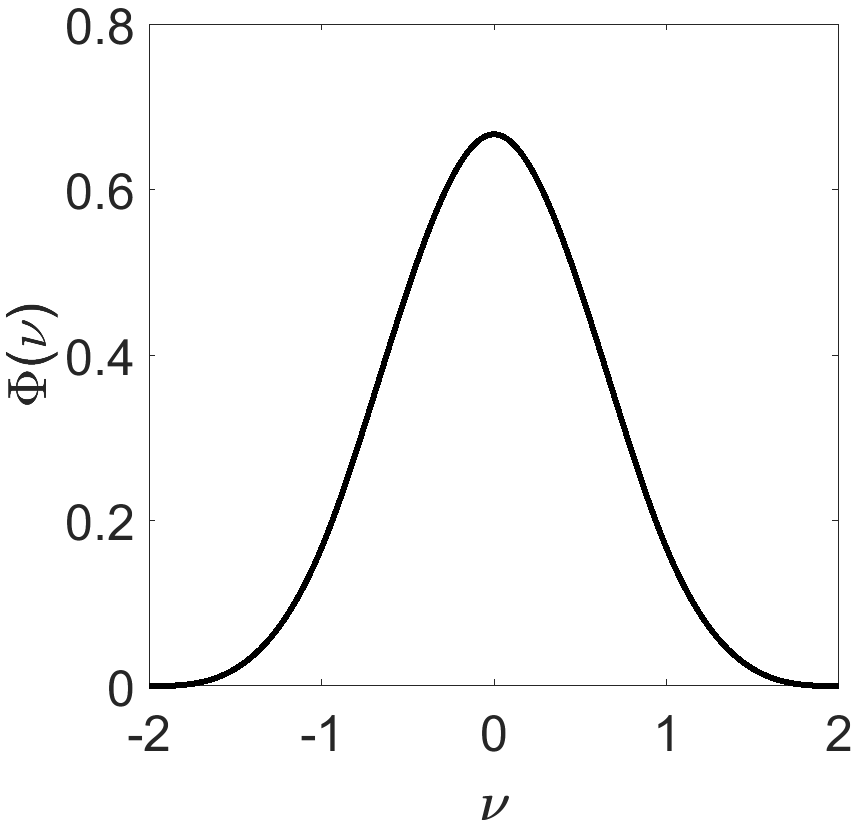}}
     \subfigure[]{
    \includegraphics[width=0.48\linewidth]{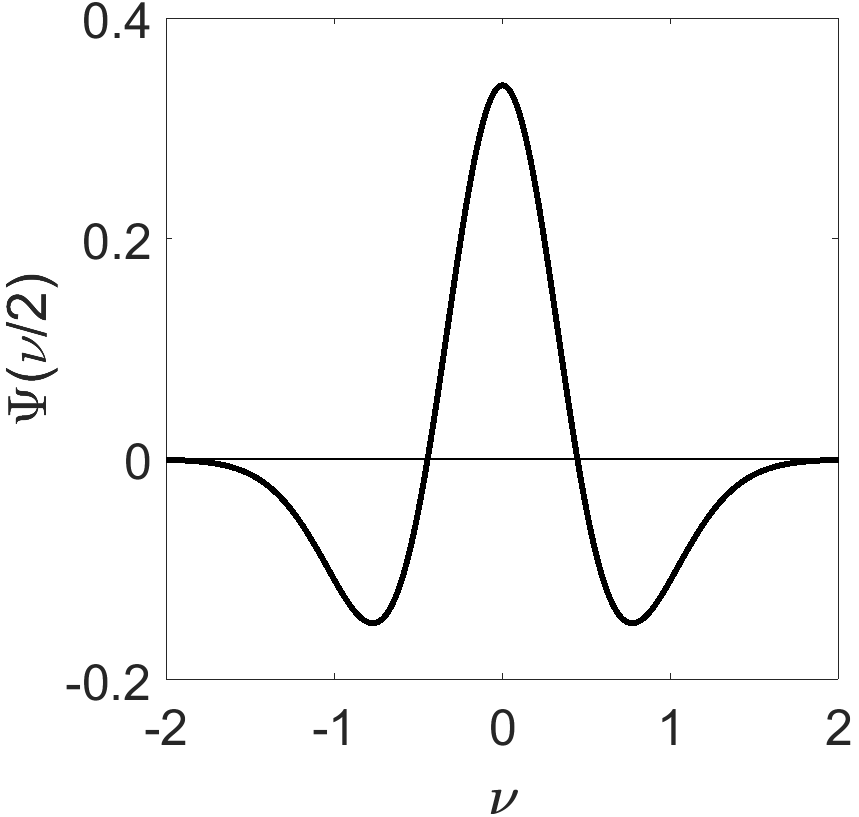}}
    \caption{(a) 1D cubic spline function $\Phi$; (b) 1D wavelet function $\Psi$.}
    \label{fig::kernel}
    \end{minipage}
\end{figure}

Since designing appropriate analysis and synthesis filter banks in image processing is a challenging task and is still open for discussion, Starck \textit{et al.}~\cite{Starck3} opted for the symmetric FIR $B_3$ filter bank.
The one dimensional (1D) cubic spline function $\Phi$ [Fig. \ref{fig::kernel}(a)] is defined~\cite{Starck3} as:
\begin{equation} \label{eq::cubic}
\Phi(\nu)=\frac{1}{12}\big(|\nu-2|^3-4|\nu-1|^3+6|\nu|^3-4|\nu+1|^3+|\nu+2|^3\big).
\end{equation}
The related filter of the scaling function is $h^{(1)}=[1,4,6,4,1]/16$ and its 2D kernel is separable and obtained by convolving two 1D cubic kernels in the $x$ and $y$ directions respectively. 
Separability allows fast computation especially for large images. Other upscaled filters $h^{(j)}$, $j\in\{2,...,J\}$, are obtained via inserting `$2^{j-1}-1$' zeros between each pair of adjacent elements in $h^{(1)}$. 
The difference between two successive resolutions of the cubic function $\Phi$ yields the wavelet function $\Psi$ [Fig. \ref{fig::kernel}(b)] as:
\begin{equation} \label{eq::wavelet}
\frac{1}{2}\Psi(\frac{\nu}{2})=\Phi(\nu) -\frac{1}{2}\Phi(\frac{\nu}{2}).
\end{equation}

\subsection{FFD Multiscale Architecture} \label{sec::ffdarchitecture}
Our findings in Sections~\ref{sec::uwtlog} and~\ref{sec::blussingrate} concern edge detection and could be used in general image processing. Here, we utilize these concepts to build a multiscale pyramid suitable for keypoint detection.

\begin{figure}
\begin{center}
\begin{minipage}[h]{1.0\linewidth}
\unitlength .5cm
    \centering
    \includegraphics[width=\linewidth]{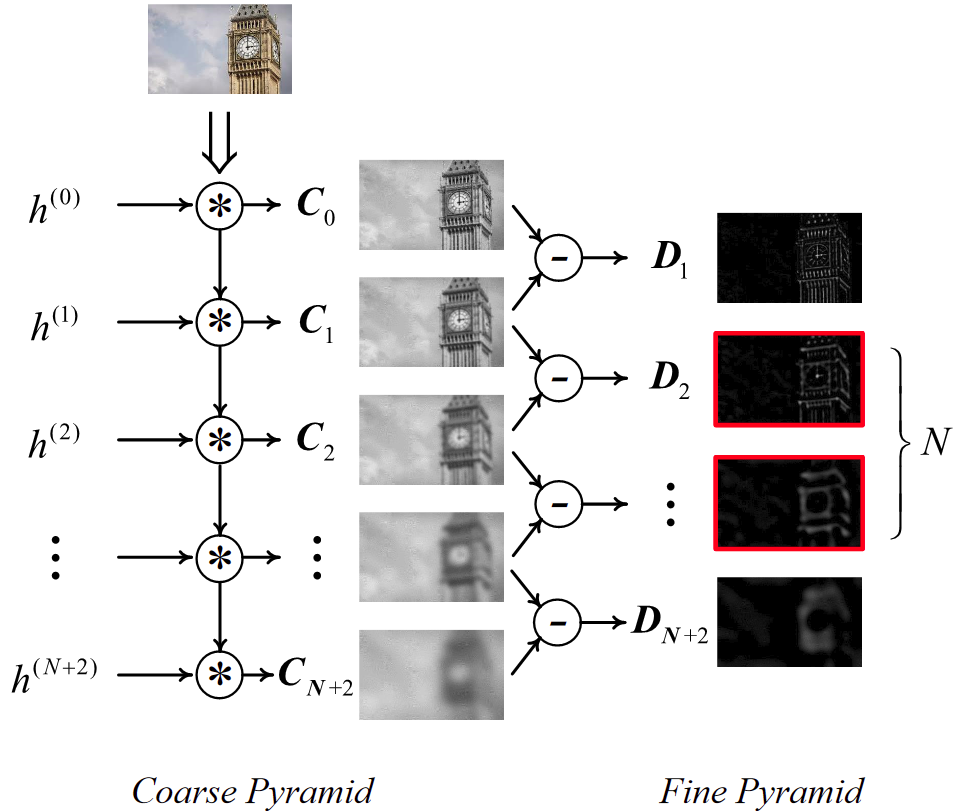}
    \caption{The framework of the proposed multiscale pyramid. $N$ fine images are output and used for feature detection in Section~\ref{sec::FeatureDetection}.}
    \label{fig::FFDframeworkscalespace}
    \end{minipage}
    \vspace{0.00mm}
\end{center}
\end{figure}

The framework of the proposed multiscale pyramid is shown in Fig.~\ref{fig::FFDframeworkscalespace}.
The input image is preliminarily smoothed by the Gaussian function with standard deviation of `$\sigma_0$' to decrease noise and other artefacts. 
The possible values for $\sigma_0$ start from 0.5, where 0.5 is the minimum value to prevent significant aliasing.
As will be discussed later, the value of $\sigma_0$ is set to 0.6 in our design.
Next, we apply the cubic spline kernel set \{$h^{(1)}$, $h^{(2)}$, ..., $h^{(N+2)}$\} to the blurred image that yields `$N+3$' coarse images. 
According to Eq. (\ref{eq1}), the smoothed input image with $\sigma_0$ is blurred by $h^{(1)}$ to yield the second coarse image; the resultant image is then convolved with $h^{(2)}$ to form the next coarse image, and so forth. 
In fact, the coarse image in the third scale level is equivalent to the convolution of the blurred input image with kernel set \{$h^{(1)}$, $h^{(2)}$\}; and likewise, the fourth coarse scale-space's image is equivalent to the convolution of the blurred input image with kernel set \{$h^{(1)}$, $h^{(2)}$, $h^{(3)}$\}, and so forth. 
This is summarized in Table \ref{table::sigmaspline}, where the sigma of the first kernel, $\sigma_1$, is equal to 1.05 and this value is approximately doubled at each scale level. 
After organizing the coarse scale-space pyramid, the next step is to form the fine scale-space pyramid. 
To this end, according to Eq. (\ref{eq2}), any two adjacent blurred images at the coarse scale-space pyramid are subtracted to yield `$N+2$' fine ones or equivalently `$N$' comparable fine ones.

As mentioned in Section \ref{sec::blussingrate}, the goal is to design a DoG kernel with $\sigma_0$ and $\mu$ of 0.627 and 2, respectively. 
On the other hand, Table \ref{table::sigmaspline} states that $\mu$ is not a constant of 2 and the sigma value of the first kernel, $\sigma_1$, is not equal to $0.627$. 
Thus, we convolve the input image with the Gaussian function with a sigma value of $\sigma_0$ in such a way that its value is around $0.627$ and, simultaneously, it yields $\mu$ of almost 2. Given $\sigma_0=\gamma\sigma_1$, where `$\gamma$' is a positive constant and $\sigma_1$ is 1.05 as shown in Table \ref{table::sigmaspline}. 
If we arrange the scale-ratios between any two consecutive coarse images into a vector `$\mathcal{M}=[\mu_{1},...,\mu_{N+2}]$' as\footnote[1]{`$N+3$' images in the coarse multiscale space pyramid yield `$N+2$' pairs of consecutive images in the fine multiscale space pyramid. 
When two Gaussian functions with sigma values $\sigma_a$ and $\sigma_b$ are convolved, the sigma value of the resultant convolution is equal to $\sqrt{\sigma_a^2+\sigma_b^2}$.}
\begin{equation} \label{eq::sr}
\mathcal{M} = \bigg[\sqrt{\frac{\sigma_0^2+\sigma_1^2}{\sigma_0^2}}, \sqrt{\frac{\sigma_0^2+\sigma_2^2}{\sigma_0^2+\sigma_1^2}}, ..., \sqrt{\frac{\sigma_0^2+\sigma_{N+1}^2}{\sigma_0^2+\sigma_{N}^2}}\bigg],
\end{equation}
then its length is `$N+2$'. In our experiment, all the elements of vector $\mathcal{M}$ approach 2 when $\gamma$ is set to 0.57. For instance, vector $\mathcal{M}$ for `$N=3$' is
\begin{equation} \label{eq::srN3}
\mathcal{M}\approx \big[2.02, 1.98, 1.99, 1.99, 1.99\big].
\end{equation}
\noindent Setting $\gamma=0.57$ yields $\sigma_0=0.6$ and its corresponding smoothing filter is: $h^{(0)}= [0.002566, 0.1655, 0.6638, 0.1655, 0.002566]$.

\begin{figure}
\begin{minipage}[h]{1.0\linewidth}
\unitlength .5cm
    \centering
    \subfigure[1st fine scale-space response]{
    \includegraphics[width=0.85\linewidth] {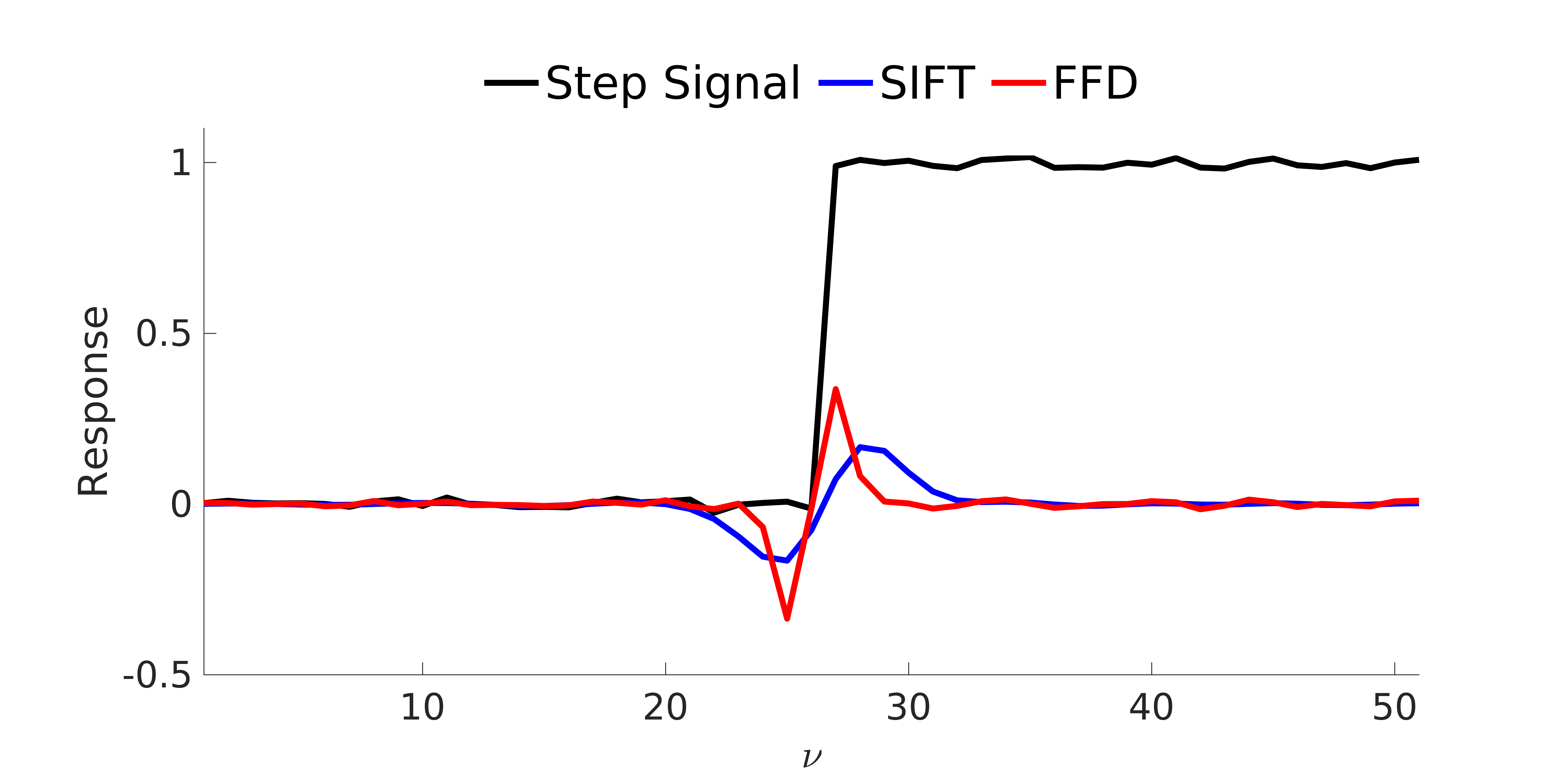}}
    \subfigure[2nd fine scale-space response]{
    \includegraphics[width=0.85\linewidth] {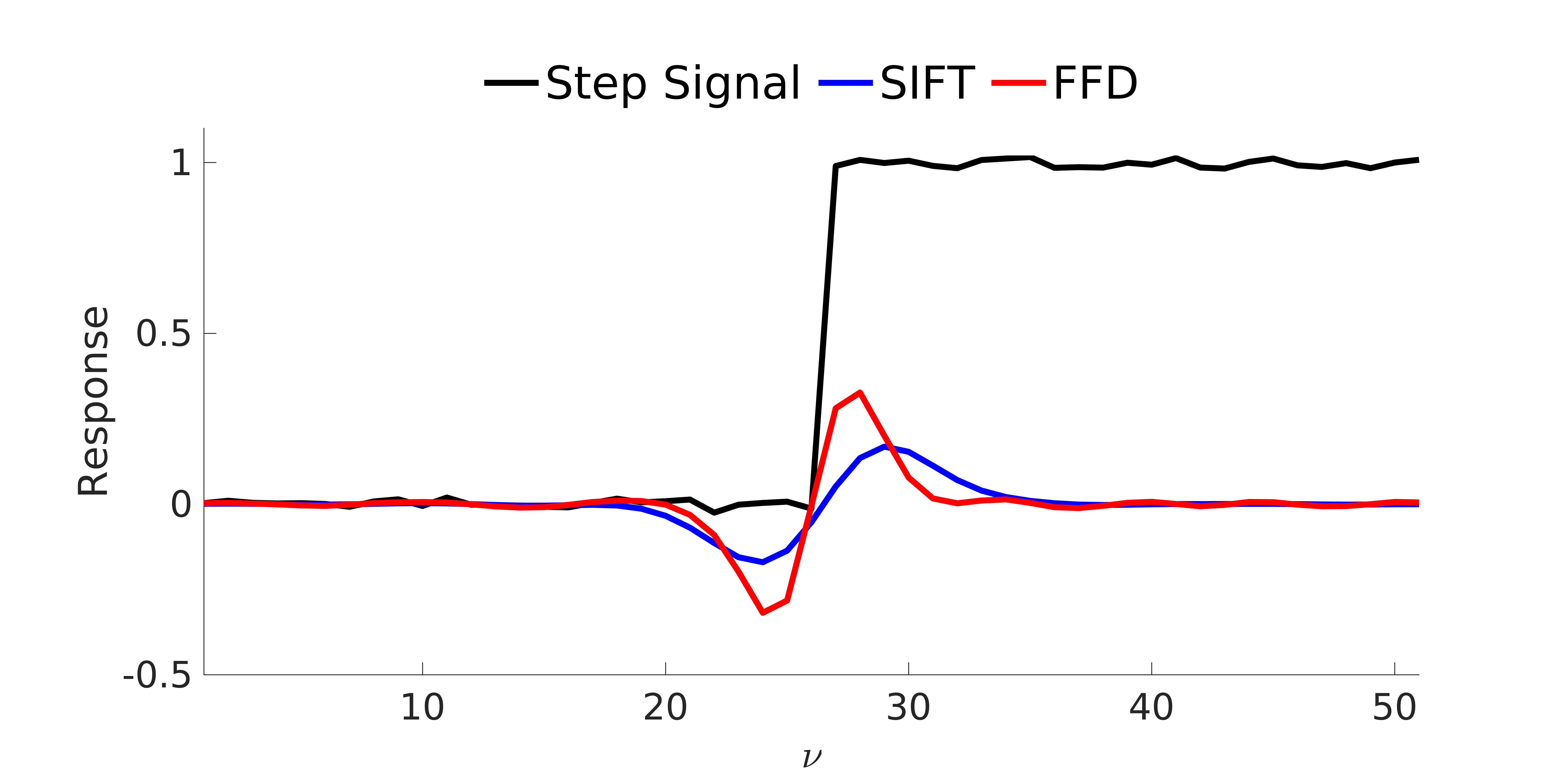}}
    \subfigure[ 3rd fine scale-space response]{
    \includegraphics[width=0.85\linewidth] {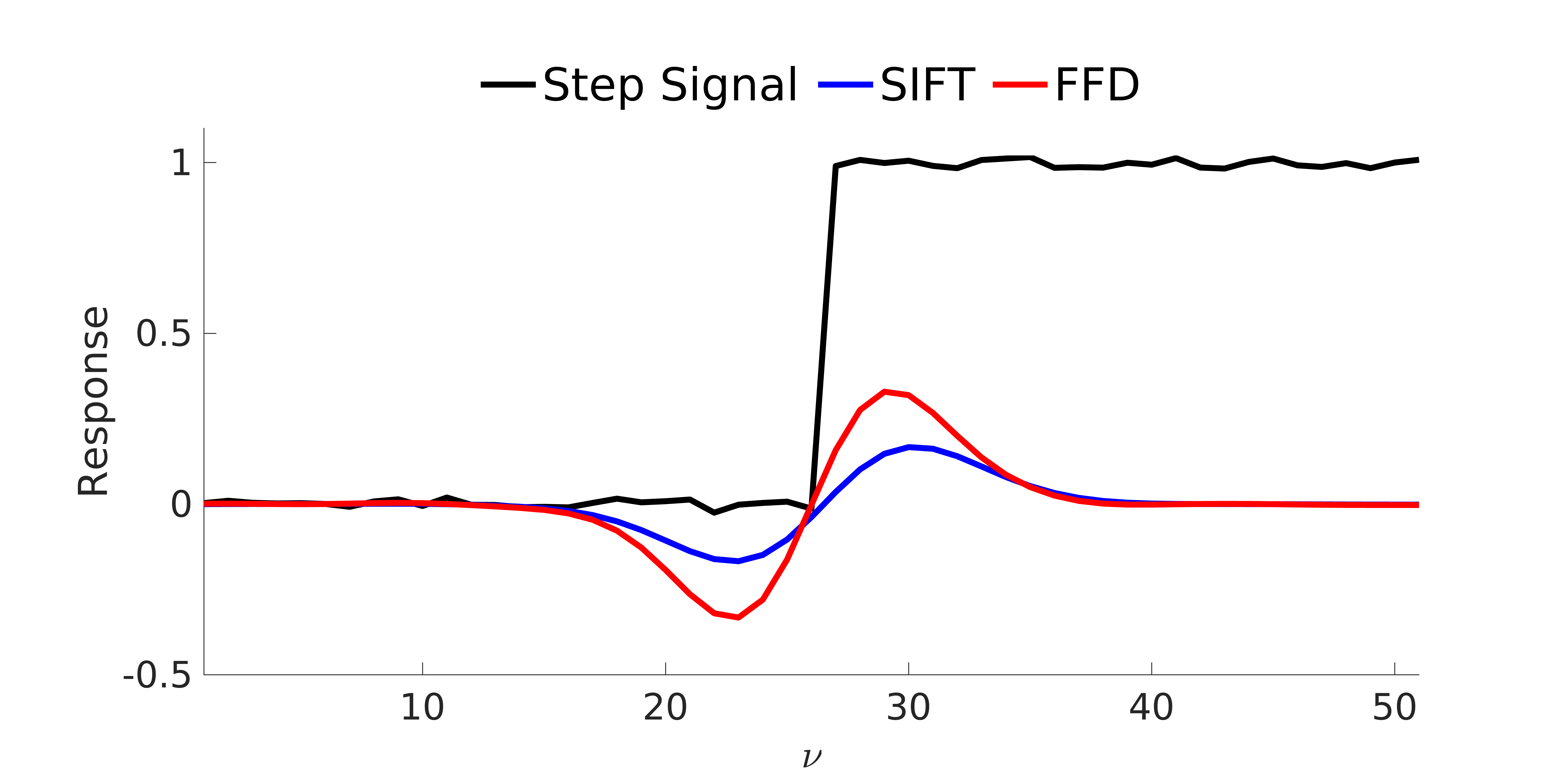}}
    \subfigure[ 4th fine scale-space response]{
    \includegraphics[width=0.85\linewidth] {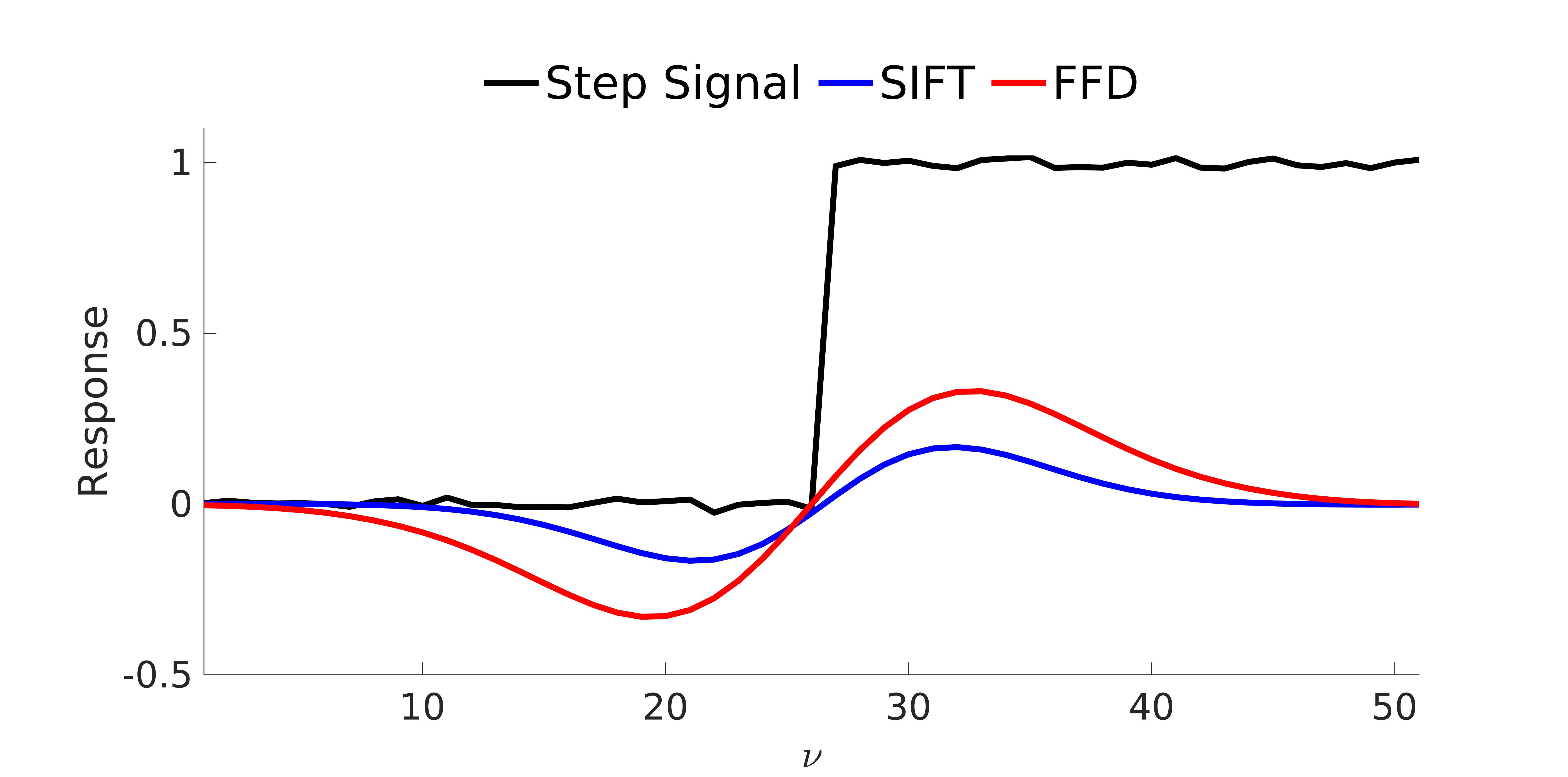}} 
    \caption{The responses of FFD and SIFT kernels to the step function.}
    \label{fig::compare}
    \end{minipage}
\end{figure}

Unlike the conventional detectors whose scale-space pyramids consist of several octaves and each octave includes some scale levels, the coarse scale-space pyramid of our feature detector contains just `$N+3$' undecimated scale levels. 
In fact, instead of downsampling the image, the kernel is upscaled. 
This feature helps us improve the localization of detected keypoints. 
To better illustrate this fact, we compare the fine scale-space responses of SIFT and FFD for the 1D step function subject to 1\% random Gaussian noise in Fig. \ref{fig::compare}. 
Parameters `$N$' in FFD and `$S$' in SIFT were set to 2, where the sigma values for SIFT and FFD are in the intervals of [1.6, 6.4] and [1.05, 9.5], respectively. 
An optimal detector should be able to detect all potential real edges and discard noisy or distorted ones. 
From the edge detection point of view, Fig. \ref{fig::compare} shows that both the methods produce smooth responses in the jagged regions contaminated with noise while in the edge area, FFD provides much stronger responses than SIFT. 
Because of smoothness, SIFT ignores some potentially reliable edges, and this Achilles heel of SIFT is more observable in the images whose texture regions are not highly discriminable like those captured at night.

Good localization of the detected feature points is another important property of a good feature detector. 
This feature plays a pivotal role in accurately estimating parameters of interest like the fundamental matrix, homography matrix, affine transform, etc. 
The location of the detected edge should be as close to the true one as possible; and in the best case, the detector should return just one point for each true edge point. 
From Fig. \ref{fig::compare}, it can be seen that FFD responses are much closer to the true edge than SIFT's, especially in the beginning levels of scale. 
As mentioned earlier, FFD covers a large interval of sigma values compared to that of SIFT for the same number of scale levels.
Adopting upsampling operators and excluding downsampling operators is at the root of good localization of the keypoints detected by FFD without any ambiguity due to interpolation. 

\begin{table}[!t]
    \centering
    \begin{minipage}[h]{1.0\linewidth}
    \centering
    \caption{The relation between kernels at different scales. $h_1$ is the cubic spline kernel ($h^{(1)}=[1,4,6,4,1]/16$).}
    \label{table::sigmaspline}
    \begin{tabular}{lcc} \toprule
    {Kernel} &{Figure} & {Sigma value}\\ \midrule
    $H_1=h^{(1)}$ &
    \begin{minipage}{.3\textwidth}
       \includegraphics[width = 1.05in,height=1.0in]{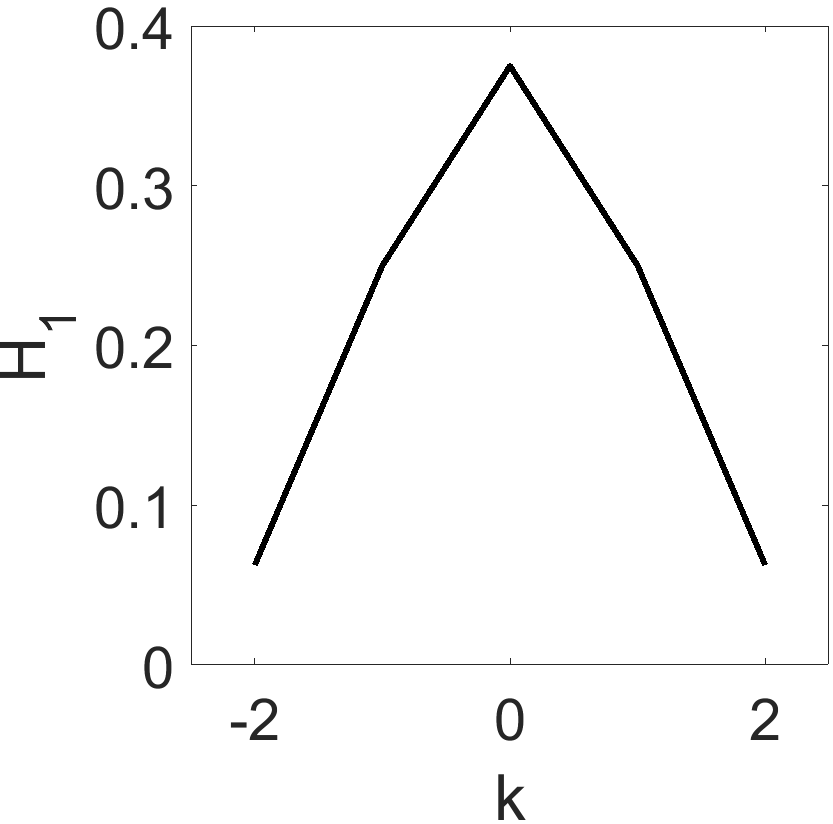}
    \end{minipage}& $\sigma_1=1.05$\\
    $H_2=h^{(2)}*h^{(1)}$ &
    \begin{minipage}{.3\textwidth}
      \includegraphics[width = 1.0in,height=1.0in]{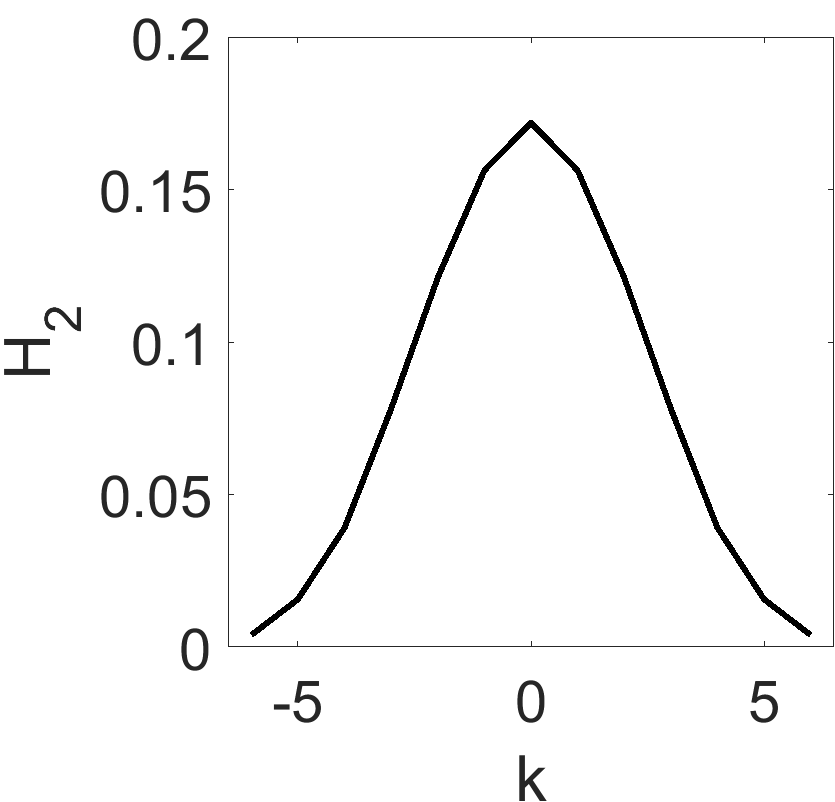}
    \end{minipage}& $\sigma_2=2.32$\\ 
    \multirow{2}{*}{\parbox{1.5cm}{$H_3=h^{(3)}*h^{(2)}*h^{(1)}$ }} &
    \begin{minipage}{.3\textwidth}
      \includegraphics[width = 1.0in,height=0.9in]{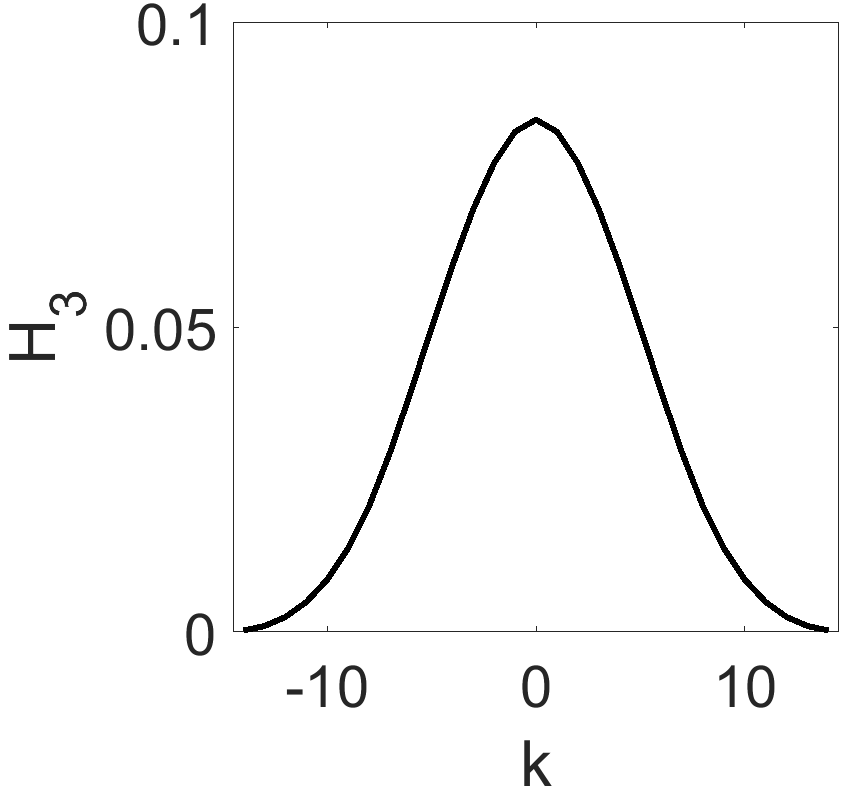}
    \end{minipage}&$\sigma_3=4.75$\\
    \multirow{2}{*}{\parbox{2cm}{$H_4= h^{(4)}*h^{(3)}*h^{(2)}*h^{(1)}$}}&
    \begin{minipage}{.3\textwidth}
      \includegraphics[width = 1.0in,height=0.9in]{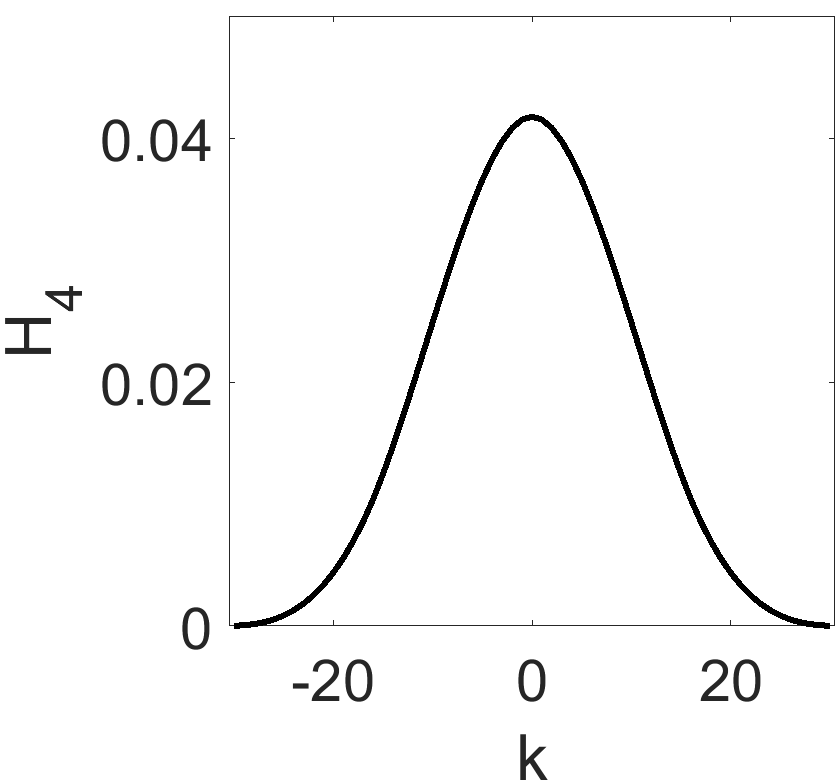}
    \end{minipage}&$\sigma_4=9.5$\\ 
    \bottomrule
    \end{tabular}
    \end{minipage}
    \vspace{0.00mm}
\end{table}
\enlargethispage{12pt}

\subsection {Feature Detection and Refinement}\label{sec::FeatureDetection}
\begin{figure*}
\begin{center}
\begin{minipage}[h]{1.0\linewidth}
\unitlength .5cm
    \centering
    \includegraphics[width=0.9\linewidth]{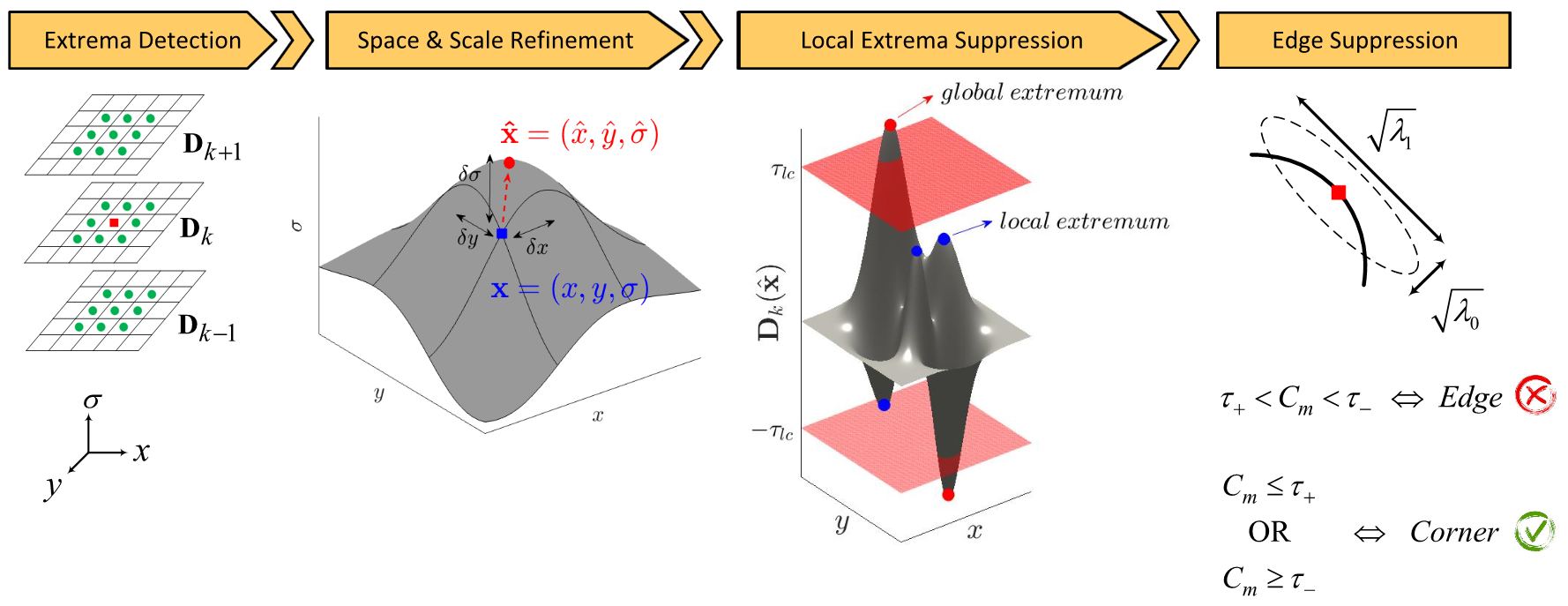}
    \caption{The framework of salient feature detection in FFD.}
    \label{fig::FFDframeworkdetection}
    \end{minipage}
    \vspace{0.00mm}
\end{center}
\end{figure*}

Hereafter, our task is to detect keypoints in the fine scale-space pyramid and refine their locations. 
Fig. \ref{fig::FFDframeworkdetection} illustrates the keypoint detection procedure in FFD.
Firstly, the candidate keypoints located at blobs in the scale-space domain are detected via a non-maximum suppression and their scale-space locations are refined [\textit{stage (I)}]. 
To reduce false positives, these candidates are then analyzed in hessian matrix and anisotropic metric. The blobs located at conjunctions are finally taken as reliable keypoints [\textit{stage (II)}]. 
In the following, each stage is discussed in detail.

\textit{I. Extrema Detection and Refinement}: 
Using a $3\times3\times3$ non-maximum suppression~\cite{non-maximum}, the extrema blobs across space and scale are detected. 
Due to discretization, the extrema are often situated between pixels in the space domain and planes in the scale domain; 
so, we examine whether they are valid extrema and if so, where their exact scale-space locations are. 
Similar to SIFT~\cite{sift}, this is done via applying the Taylor expansion to the extrema. 
Given that the candidate keypoint is located at $\mathbf{x}=(x,y,\sigma)$ in the $k$th fine image $\mathbf{D}_k$, $k\in\{2,...,N+1\}$. 
The quadratic Taylor expansion of the intensity $\bf D_k(\mathbf{x})$ is defined as
\begin{equation} \label{eq::taylor1}
\mathbf{D}_k(\mathbf{x+\Delta}) =\mathbf{D}_k(\mathbf{x})+ {\frac{\partial \mathbf{D}_k(\mathbf{x})}{\partial \mathbf{x}}\bf{\Delta}}+
\frac{1}{2}{\bf{\Delta}}^T \frac{\partial^2 \mathbf{D}_k(\mathbf{x})}{\partial \mathbf{x}^2}\bf{\Delta},
\end{equation}
where $\mathbf{\Delta}=(\delta x,\delta y,\delta \sigma)$ is the offset of the keypoint from the given point $\bf x$. 
Taking the derivative of Eq. (\ref{eq::taylor1}) with respect to $\mathbf{\Delta}$ and setting it to zero yields the offset of the candidate keypoint:
\begin{equation} \label{eq::taylor2}
\hat{\mathbf{\Delta}} =-\frac{\partial^2 \mathbf{D}_k^{-1}(\mathbf{x})}
{\partial \mathbf{x}^2}\frac{\partial \mathbf{D}_k(\mathbf{x})}{\partial \mathbf{x}}.
\end{equation}
The new location of the keypoint of interest will be $\mathbf{\hat x}=\mathbf{x}+\hat{\mathbf{\Delta}}$ if each element of the offset vector $\hat{\mathbf{\Delta}}$ is smaller than 0.5. Otherwise, the candidate keypoint is not a valid extremum and thus is discarded.

Detected blobs in the fine scale-space pyramid could be local or global extrema. 
Compared to global extrema, local ones have low contrast and can be removed by applying a contrast threshold, $\tau_{lc}$, to the intensity values of the extrema, i.e. $\mathbf{D}_k(\hat{\mathbf{x}})$. 
The intensity values are obtained via inserting Eq. (\ref{eq::taylor2}) into Eq. (\ref{eq::taylor1}):
\begin{equation} \label{eq::taylor3}
\mathbf{D}_k(\hat{\mathbf{x}}) =\mathbf{D}_k(\mathbf{x}) +\frac{1}{2}\frac{\partial \mathbf{D}_k(\mathbf{x})}{\partial \mathbf{x}}\hat{\mathbf{\Delta}}.
\end{equation}
In practical image processing, the extrema with high contrast are more favourable. Bear in mind that Eq.~(\ref{eq::taylor3}) could not remove superimposed blobs since they have large amplified values [see Fig.~\ref{fig::dwd_d}].
In the case of destructive superimposition, this equation may also discard potential keypoints, increasing false negative.

\textit{II. Edge Suppression}: As the detected blobs may not be reliable, the goal is to select the reliable ones, located at conjunctions. 
To this end, we use the anisotropy definition proposed by~\cite{OptimalOrientation}. Let's define the tensor $\bf{J}$ of the keypoint at $\mathbf{x}=(x,y,\sigma)$ as follows:
\begin{equation} \label{eq::tensor}
\bf{J} = 
\begin{bmatrix}
    \frac{\partial^2 \mathbf{D}_k(\mathbf{x})}{\partial x^2}
    & \frac{\partial^2 \mathbf{D}_k(\mathbf{x})}{\partial x \partial y}\\[0.3em]
    \frac{\partial^2 \mathbf{D}_k(\mathbf{x})}{\partial x \partial y} & \frac{\partial^2 \mathbf{D}_k(\mathbf{x})}{\partial y^2}
\end{bmatrix}=
\begin{bmatrix}
    \mathbf{J}_{xx} & \mathbf{J}_{xy}\\[0.3em]
    \mathbf{J}_{xy} & \mathbf{J}_{yy}
\end{bmatrix}
.
\end{equation}
Then anisotropy parameter $C_m$ for the given pixel is defined as
\begin{equation} \label{eq::anisotropy}
C_m = \bigg(\frac{\lambda_1-\lambda_0}{\lambda_1+\lambda_0}\bigg)^c,
\end{equation}
where $\lambda_0$ and $\lambda_1$ are the two eigenvalues of Eq. (\ref{eq::tensor}), and `$c$' is a positive constant. 
In order to avoid negative values, $c$ is chosen as 2. 
For a keypoint located at conjunction, we have $\lambda_1 \approx \lambda_0$ and subsequently $C_m \approx 0$.
According to the definition, they are computed from the following equation
\begin{equation} \label{eq::eigen}
\lambda_{0,1} = \frac{1}{2}\bigg(\mathbf{J}_{yy}+ \mathbf{J}_{xx}\pm \sqrt{\big(\mathbf{J}_{yy}-\mathbf{J}_{xx})^2+4\mathbf{J}^2_{xy}}\bigg).
\end{equation}
Inserting Eq. (\ref{eq::eigen}) into the anisotropy definition, i.e. Eq. (\ref{eq::anisotropy}), yields 
\begin{equation} \label{eq::anisotropyfinal}
C_m = \bigg(\frac{\sqrt{\big(\mathbf{J}_{yy}-\mathbf{J}_{xx})^2+4\mathbf{J}^2_{xy}}}{\mathbf{J}_{yy}+\mathbf{J}_{xx}}\bigg)^2 = \\
1-4\frac{Det(\mathbf{J})}{Tr^2(\mathbf{J})},
\end{equation} 
where $Det(\mathbf{J})$ and $Tr(\mathbf{J})$ denote the determinant and the trace of the tensor $\bf{J}$ in Eq. (\ref{eq::tensor}):
\begin{equation} \label{eq::det}
Det(\mathbf{J})=\mathbf{J}_{xx}\mathbf{J}_{yy}-\mathbf{J}^2_{xy},
\end{equation}
and
\begin{equation} \label{eq::tr}
Tr(\mathbf{J})=\mathbf{J}_{xx}+\mathbf{J}_{yy}.
\end{equation}
$C_m$ in Eq. (\ref{eq::anisotropyfinal}) takes values in the interval of [0, 1] as $Tr(\mathbf J)\geq2\sqrt{Det(\mathbf J)}$. 
If the determinant of $\mathbf{J}$ has a large positive value, its eigenvalues are large and, subsequently, we have strong edges at multiple orientations such as conjunctions and corners. 
In practice, the determinant can also take negative values. 
To facilitate analysis, we rewrite Eq. (\ref{eq::eigen}) based on the determinant $Det(\mathbf{J})$ and the trace $Tr(\mathbf{J})$ of the tensor $\bf{J}$ as:
\begin{equation} \label{eq::rephrase}
\lambda_{0,1} = \frac{1}{2}\bigg(Tr(\mathbf{J})\pm \sqrt{Tr^2(\mathbf{J})-4Det(\mathbf{J})}\bigg) = \frac{1}{2}(\alpha\pm\beta),
\end{equation}
where $\alpha=Tr(\mathbf{J})$ and $\beta=\sqrt{Tr^2(\mathbf{J})-4Det(\mathbf{J})}=\sqrt{\alpha^2-4Det(\mathbf{J})}$. If the determinant $Det(\mathbf{J})$ takes a negative value, $\beta$ is then greater than $\alpha$ and this means that the two eigenvalues have opposite signs. Similar to the large positive response, a large negative response also indicates the presence of multiple edges like saddle points~\cite{NegDet,Lindeberg3}. Thus, we define two predetermined anisotropy thresholds, one for the positive determinant $\tau_{+}$ and the other for the negative determinant $\tau_{-}$. For each candidate keypoint, the anisotropy metric $C_m$ is calculated via Eq. (\ref{eq::anisotropyfinal}) and if it meets the predetermined thresholds, i.e. $C_m \leq\tau_{+}$ or $C_m \geq\tau_{-}$, then it is located at conjunction and is thus labelled as a reliable keypoint; otherwise, it is considered as an edge response and discarded. 

\section{Experimental Results}
\label{sec::experimental}
We evaluate our proposed FFD against several state-of-the-art ones, including
\begin{itemize}
    \item Multiscale methods: SIFT~\cite{sift}, SURF~\cite{surf}, KAZE~\cite{kaze}, BRISK~\cite{brisk} and HarrisZ~\cite{harrisz}; 
    \item Learning methods: TILDE~\cite{deepnet1}, DNet~\cite{detnet}, TCDET~\cite{detdeep}, LIFT~\cite{lift}, SuperPoint~\cite{superpoint} and D2Net~\cite{d2net}.
\end{itemize}
\begin{figure*}
\begin{center}
\centering
    \begin{minipage}[h]{1.0\linewidth}
    \centering
    \subfigure[Hannover]{
    \includegraphics[width=0.48\linewidth]{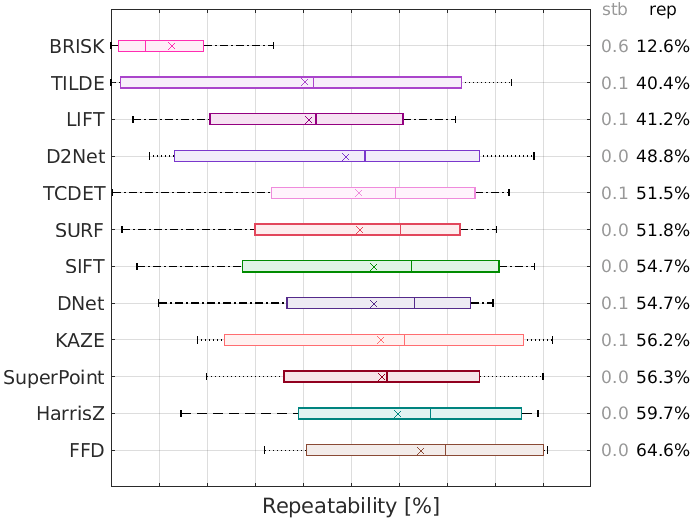}}
    \subfigure[WebCam]{
    \includegraphics[width=0.48\linewidth]{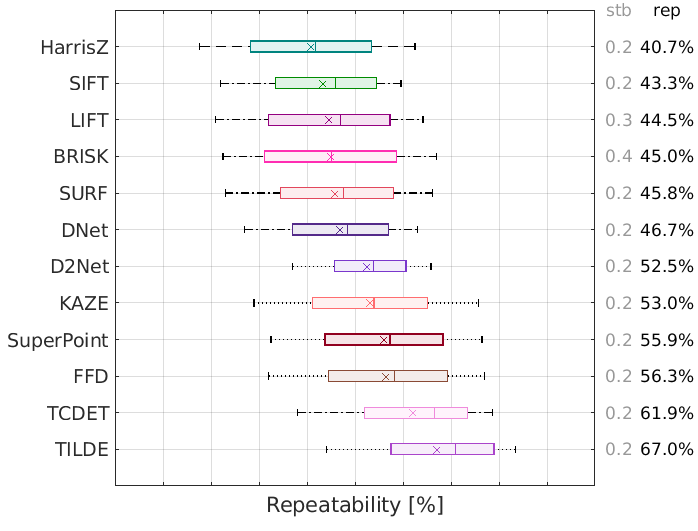}
    \label{fig::WebC}}
    \end{minipage}
    \vspace{0.00mm}
    \begin{minipage}[h]{1.0\linewidth}
    \centering
    \subfigure[VGG Affine]{
    \includegraphics[width=0.48\linewidth]{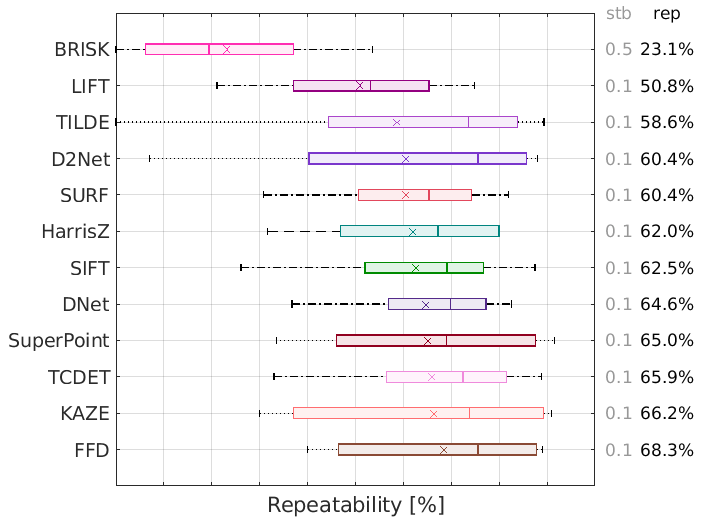}}
    \subfigure[Edge Foci]{
    \includegraphics[width=0.48\linewidth]{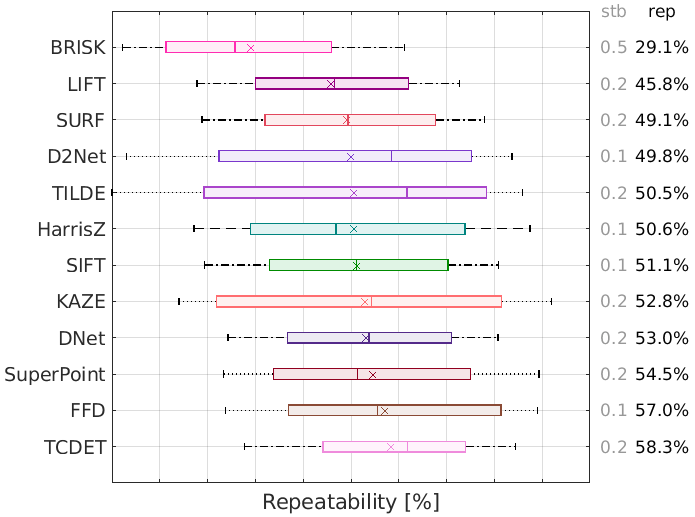}}
    \end{minipage}
    \vspace{0.00mm}
    \begin{minipage}[h]{1.0\linewidth}
    \centering
    \subfigure[HSequences, Illumination]{
    \includegraphics[width=0.48\linewidth]{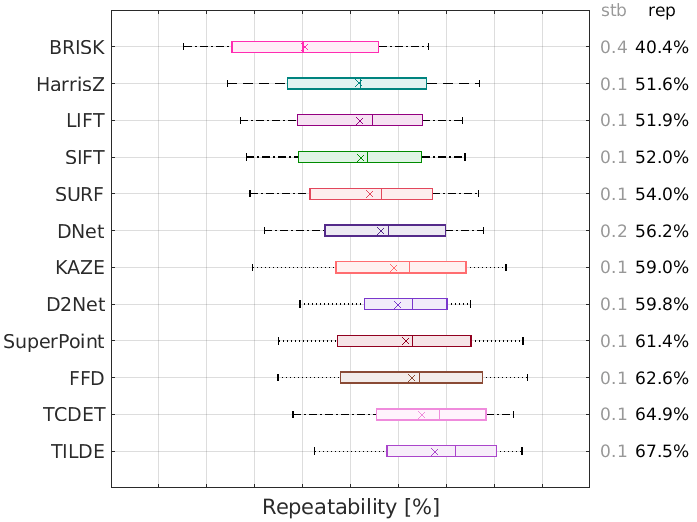}
    \label{fig::hpi}} 
    \subfigure[HSequences, Viewpoint]{
    \includegraphics[width=0.48\linewidth]{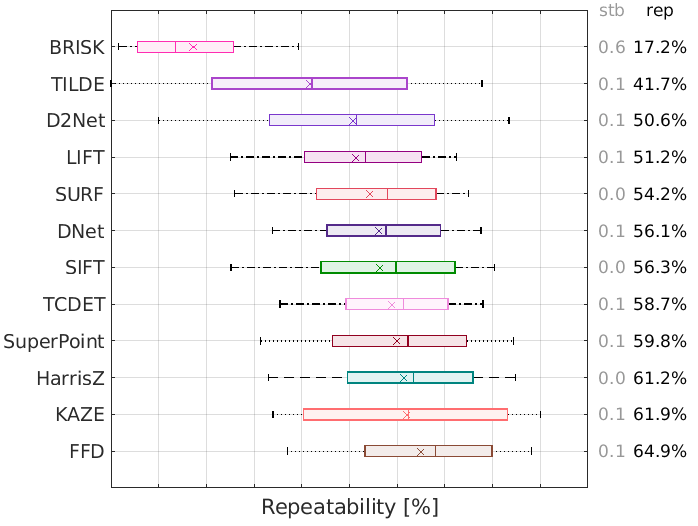}\label{fig::hpv}}
    \end{minipage}
    \vspace{0.00mm}
    \caption{Repeatability of different feature detectors over Hannover, Webcam, VGG Affine, Edge Foci, and HSequences (Illumination and Viewpoint) databases respectively.}
    \label{fig::results}
\end{center}
\end{figure*}
We used their implementations from OpenCV\footnote{\url{https://opencv.org/}} except HarrisZ and the learning detectors.
The code of HarrisZ is available in~\cite{harrisz}. The codes and pre-trained models of the learning methods released by the authors were used here. 
In general, KAZE provides better results than its accelerated variant and this motivated us to compare our detector with that. It is worth noting that the computational time of AKAZE is also reported in the computational time section. 
The number of scales per octave for multiscale feature detectors were set to 3. 
In order to provide sufficient keypoints for each image, we set the detection thresholds in SIFT, KAZE and SuperPoint to 0.025, 0.0003 and 0.001, respectively; similarly, the corner detection threshold in BRISK and the keypoint detection threshold in SURF were set to 15 and 300, respectively.
We used the default values for other parameters and a maximum of 10$k$ best keypoints per image of each feature detector were selected. 

For FFD\footnote{\url{https://github.com/mogvision/FFD}}, parameters $N$ and $\tau_{lc}$ were set to 3 and 0.05, respectively. A blob is labelled as an edge response if $0.7 \leq C_m \leq 1.5$, where the boundaries between edge and corners, i.e. $\tau_{+}$ and $\tau_{-}$, were set to $0.7$ and $1.5$, respectively.
The feature points are assessed by repeatability \& stability, robustness, visual localization, 3D reconstruction, golden parameter values, keypoint distribution and computational time, detailed in the following sections.
\subsection{Repeatability and Stability with Homography Datasets} \label{sec::Repeatability}
Here we validate the performance of the local feature detectors in repeatability and instability in the pipeline developed by Lenc and Vedaldi~\cite{pipeline}. This pipeline was applied to several publicly available homography databases including Hannover~\cite{Hannover}, Webcam~\cite{Webcam}, VGG Affine~\cite{survey2005b}, Edge Foci~\cite{Foci}, and HSequences~\cite{HPatch}.
Mikolajczyk \textit{et al.}~\cite{survey2005b} define the repeatability score as the fraction of keypoints that match between images with sufficient geometric overlap up to the ground-truth homography matrix. 
But it is revised by Lenc and Vedaldi~\cite{pipeline} through normalization. 
They also introduce the instability score that quantifies the stability of the detectors across different thresholds. According to its definition, the instability of a feature detector is calculated as the standard deviation of the repeatability scores, which is then normalized by the average repeatability.

Figure \ref{fig::results} shows the box percentiles (first and third quartile) and the whisker percentiles of results of different feature detectors (10\% and 90\%).
In the databases containing illumination changes, i.e. HSequences-illumination Fig.~\ref{fig::hpi} and WebCam Fig.~\ref{fig::WebC}, the learning detectors often yield higher repeatability than the traditional ones due to their pre-training with data augmentations.
In this experiment, TILDE performs well and the proposed FFD is also competitive with the learning detectors, especially in the third quartile and median values.
However, TILDE is not affine invariant. In the presence of viewpoint changes, TCDET and SuperPoint outperform other learning-based feature detectors.
FFD gains the highest repeatability score in three out of four viewpoint databases and TCDET wins over the remaining one. 
KAZE tends to have high repeatability, indicating that it is affine invariant.
The stability error of most feature detectors is less than 10\% while  BRISK has the largest variation.
If we consider the results of feature detectors over both the illumination and viewpoint sequences, it can be concluded that FFD, SuperPoint and KAZE achieve the best performance.

\subsection{Robustness of Feature Detectors}
\begin{figure}
\begin{center}
\centering
\unitlength .5cm
    \begin{minipage}[h]{1.0\linewidth}
    \centering
    \includegraphics[width=0.7\linewidth]{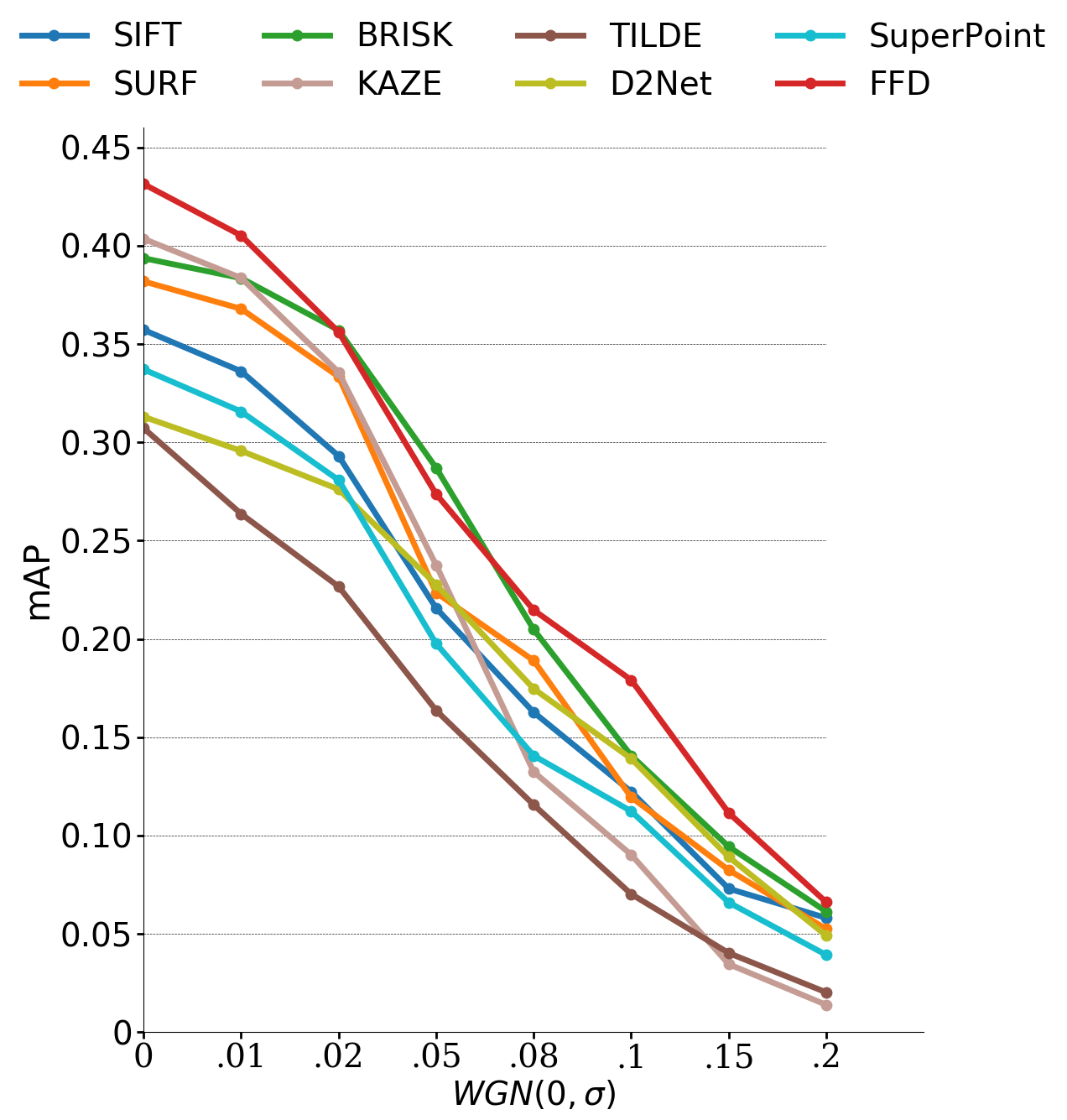}
    \label{fig::PR_noise}
    \end{minipage}
    \caption{mAP results of different feature detectors over the  images in the homography databases corrupted by the additive white-Gaussian noise (WGN) with a standard deviation from 0.01 to 0.2.}
    \label{fig::homo_noise}
\end{center}
\end{figure}

Here we evaluate the robustness of the feature detectors against noise and blurring. The experiments were run over the homography databases summarised in Section~\ref{sec::Repeatability}, and the detected keypoints were assessed by mean average precision (mAP). 
The additive white Gaussian noise (WGN) with a standard deviation from 0.01 to 0.2 was added to the images, even though noise can be space-variant in practical imaging. The results on the synthesized data are reported in Fig.~\ref{fig::homo_noise}, where FFD and BRISK show more resistance against noise than the other methods.\\
The images were also blurred by an averaging filter with various kernel sizes from $3\times3$ to $13\times13$. 
The results are reported in Table~\ref{table::homo_blur}. Since the number of detected keypoints is affected by blurring, we also reported the number of established correspondences. 
In terms of the number of correspondences, D2Net is less affected by blurring than the others but its mAP drops more considerably. 
Taking both the metrics into account, FFD, KAZE and SuperPoint are less prone to blurring.

\begin{table*}[!t]
    \begin{minipage}[h]{1.0\linewidth}
    \centering
    \caption{The mAP (in parentheses) and the number of the established correspondences of different methods for detecting keypoints from the images blurred over different sizes of kernels.}
    \label{table::homo_blur}
    \begin{tabular}{lccccccccc} 
    \toprule
     {\bf Detector}
     & {Blur-free}
        & {\bf $3\times3$}
            & {\bf $5\times5$}
                & {\bf $7\times7$}
                    & {\bf $9\times9$}
        & {\bf $11\times11$}
            & {\bf $13\times13$}\\ \midrule %
    \verb//{\it SIFT}&2000.8 (0.486)&1897.9  (0.481)&903.9  (0.489)&542.3  (0.473)&367.8  (0.463)&277.9 (0.451)&220.1 (0.414)\\
    \verb//{\it SURF}&  1515.8 (0.495)&1210.0 (0.488)&1010.6 (0.476)&767.9 (0.471)&550.3 (0.476)&401.1 (0.479)&299.9 (0.456)\\
    \verb//{\it BRISK}& 2162.6 (0.485)&1299.2 (0.468)&638.6  (0.451)&371.2 (0.499)&249.7 (0.498)&183.4 (0.445)&149.9 (0.409)\\
    \verb//{\it KAZE} &1674.0 (0.513)&1511.5 (0.491)&1261.3 (0.509)&991.8 (0.518)&753.2 (0.523)&562.7 (0.512)&421.4 (0.504)\\ %
    \verb//{\it TILDE}& 1232.9 (0.422)&1023.1 (0.421)&872.3  (0.417)&623.3 (0.409)&491.8 (0.398)&374.9 (0.386)&277.3 (0.378)\\
    \verb//{\it SuperPoint}& 1259.1 (0.451)&1061.1 (0.453)&899.3 (0.475)&851.4 (0.463)&696.4 (0.425)&657.4 (0.396)&538.2 (0.387)\\
    \verb//{\it D2Net}&2842.3 (0.437)&2484.1 (0.435)&2196 (0.421)&1858.6 (0.419)&1665.8 (0.395)&1402.6 (0.371)&1233.8  (0.365)\\ %
    \verb//{\it FFD}&1610.9 (0.548)&1423.1 (0.531)&1247.5 (0.550)&912.1  (0.559)&713.8  (0.538)&587.5  (0.513)&448.9  (0.477)\\ 

    \bottomrule
    \end{tabular}
    \end{minipage}
    \vspace{0.00mm}
\end{table*}

\subsection{Visual Localization} \label{sec::localization}
Visual localization is an important task that needs an accurate estimation of the position and orientation of the cameras. 
Real-world conditions like distortion, noise and day-night transitions severely affect the contents of the images and 
feature matching of such images is thus usually challenging. 
Aachen Day-Night dataset~\cite{sattler2018benchmarking} contains 4,328 day-time images and 98 night-time queries. The performance of local feature descriptors is evaluated by a pre-defined visual localization pipeline\footnote{\url{https://github.com/tsattler/visuallocalizationbenchmark/tree/master/local\_feature\_evaluation}}.
The results of successfully localized images are reported with three tolerances in estimation errors of position and orientation: (0.5m, 2 deg.), (1m, 5 deg.) and (5m, 10 deg.). HardNet++~\cite{hardnet++} was employed as the local feature descriptor for all the extracted keypoints. 

The numerical results are tabulated in Table~\ref{table::resultslocalization}.
The table shows that FFD  achieves the best performance over all the three defined thresholds by significant margins. For strict accuracy thresholds in the estimated localization, our technique works better than all the others by as much as 2\%, verifying its outstanding efficacy for localisation of the detected keypoints.

\begin{table}[!t]
    \begin{minipage}[h]{1.0\linewidth}
    \centering
    \caption{Evaluation results (\%) of different defectors for visual localization on the Aachen dataset. The first and second best results are highlighted in boldface and with underline, respectively.}
    \label{table::resultslocalization}
    \begin{tabular}{lccc} 
    \toprule
     {\bf Detector}
            & {\bf (0.5m, \angle{2})}
                & {\bf (1m, \angle{5})}
                    & {\bf (5m, \angle{10})}\\ \midrule %
    \verb//{\it SIFT}& \underline{42.9} & 56.1 & \underline{80.6} \\ %
    \verb//{\it SURF}& 38.8  & 55.1  & 73.5 \\ %
    \verb//{\it BRISK}& 39.8 & \underline{59.2} & 77.6 \\ %
    \verb//{\it HarrisZ}& 41.8 & 57.1 & 75.5\\ %
    \verb//{\it KAZE}& 40.6 & 53.0 & 74.4 \\~\\ 

    \verb//{\it LIFT}& 35.6 & 53.1 & 67.3 \\ %
    \verb//{\it DNet}& 37.2 & 54.1 & 68.4  \\ %
    \verb//{\it TILDE}& 38.8 & 54.1 & 69.4 \\ %
    \verb//{\it TCDET}& 39.8 & 55.1 & 72.5 \\ %
    \verb//{\it SuperPoint}& 40.8 & \underline{59.2} & 78.6 \\ %
    \verb//{\it D2Net}& 40.8 & 56.1 & 75.5 \\~\\ %
  
    \verb//{\it FFD}& \textbf{44.9} & \textbf{60.2} & \textbf{81.6}\\ 
    \bottomrule
    \end{tabular}
    \end{minipage}
    \vspace{0.00mm}
\end{table}

\subsection{3D Reconstruction} \label{sec::reconstruction}

\renewcommand{\tabcolsep}{3pt}
\begin{table*}
    \begin{minipage}[h]{1.0\linewidth}
    \centering
    \caption{Evaluation results of different feature detectors for the 3D reconstruction benchmark.}
    \label{table::results3dults}
    \begin{tabular}{lccccccc} 
    \toprule
     {\bf \small{Dataset}}
        & {\bf \small{Descriptor}}
            & {\bf \small{\# Registered}}
                & {\bf \small{\# Observations}}%
                   & {\bf \small{\# Inlier Pairs}}%
                       & {\bf \small{\# Inlier Matches}}%
                            & {\bf \small{\# Sparse Points}}
                             & {\bf \small{\# Dense Points}}\\
                           \small{(\# Images)} && & &  &  & \\ \midrule %
    \verb//{\bf \small{Herzjesu}} 
    &\verb//{\it SIFT}&8&38\kformat&28&46\kformat&11\kformat&\underline{244\kformat}\\ (8)
    &\verb//{\it BRISK}&8&39\kformat&28&38\kformat&12\kformat&239\kformat\\
    &\verb//{\it KAZE}&8&41\kformat&28& 43\kformat&13\kformat&243\kformat\\~\\
    &\verb//{\it TILDE}&8&72\kformat&28&\underline{103\kformat}&19\kformat&240\kformat\\
    &\verb//{\it SuperPoint}&8&66\kformat&28&86\kformat&18\kformat&242\kformat\\
    &\verb//{\it D2Net}&8&\underline{83\kformat}&28&91\kformat&\underline{24\kformat}&\textbf{245\kformat}\\~\\
    &\verb//{\it FFD}&8&\textbf{86\kformat} &28&\textbf{118\kformat}&\textbf{26\kformat}&\textbf{245\kformat}\\\cmidrule{1-8}
    \verb//{\bf Fountain}
    &\verb//{\it SIFT}&11&81\kformat&55&118\kformat&20\kformat&\underline{307\kformat}\\ (11)
    &\verb//{\it BRISK}&11&75\kformat&55&81\kformat&21\kformat&304\kformat\\
    &\verb//{\it KAZE}&11&67\kformat&55&75\kformat&20\kformat&304\kformat\\~\\
    &\verb//{\it TILDE}&11&101\kformat&55&\underline{169\kformat}&24\kformat&306\kformat\\
    &\verb//{\it SuperPoint}&11&103\kformat&55&155\kformat&26\kformat&305\kformat\\
    &\verb//{\it D2Net}&11&\underline{127\kformat}&55&155\kformat&\underline{33\kformat}&306\kformat\\~\\
    &\verb//{\it FFD}&11&\textbf{166\kformat}&55&\textbf{283\kformat}&\textbf{38\kformat} &\textbf{308\kformat}\\\cmidrule{1-8}
    \verb//{\bf \small{Madrid}}
    &\verb//{\it SIFT}&743&1.26\mformat&896\kformat&68.8\mformat&251\kformat&1.18\mformat\\{\bf \small{Metropolis}}
    &\verb//{\it BRISK}&731&1.19\mformat&897\kformat&64.7\mformat&237\kformat&1.16\mformat\\(1,344)
    &\verb//{\it KAZE}&\underline{784}&1.33\mformat&\underline{898}\kformat&\underline{70\mformat}&\underline{274\kformat}&\underline{1.31\mformat}\\~\\
    &\verb//{\it TILDE}&635&696\kformat&887\kformat&48.2\mformat&164\kformat&1.05\mformat\\
    &\verb//{\it SuperPoint}&723&867\kformat&897\kformat&56.9\mformat&173\kformat&1.15\mformat\\
    &\verb//{\it D2Net}&758&\textbf{1.52\mformat}&\underline{898\kformat}&66.2\mformat&264\kformat&1.26\mformat\\~\\
    &\verb//{\it FFD}&\textbf{813}&\underline{1.43\mformat}&\textbf{899\kformat}&\textbf{73.1\mformat}&\textbf{315}\kformat&\textbf{1.36\mformat}\\\cmidrule{1-8}
    \verb//{\bf \small{Gendarmen-}}
    &\verb//{\it SIFT}&\underline{1188}&2.55\mformat&1.066\mformat&88.4\mformat&472\kformat&3.04\mformat\\{\bf \small{markt}}
    &\verb//{\it BRISK}&1145&2.36\mformat&1.065\mformat&74.9\mformat&412\kformat&3.01\mformat\\(1,463)
    &\verb//{\it KAZE}&1180&2.71\mformat&\textbf{1.069\mformat}&\textbf{93.6\mformat}&563\kformat&3.06\mformat\\~\\
    &\verb//{\it TILDE}&1083&2.05\mformat&1.051\mformat&58.8\mformat&326\kformat&2.98\mformat\\
    &\verb//{\it SuperPoint}&1132&1.84\mformat&\underline{1.067\mformat}&64.8\mformat&356\kformat&\underline{3.14\mformat}\\
    &\verb//{\it D2Net}&1154&\underline{2.82\mformat}&\underline{1.067\mformat}&90.3\mformat&\underline{611\kformat}&3.08\mformat\\~\\
    &\verb//{\it FFD}&\textbf{1216}&\textbf{2.96\mformat}&\textbf{1.069\mformat}&\underline{92.4\mformat}&\textbf{635\kformat}&\textbf{3.23\mformat}\\\cmidrule{1-8}
    \verb//{\bf \small{Tower of}}
    &\verb//{\it SIFT}&\underline{1126}&\underline{3.19\mformat}&\underline{1.238\mformat}&113.4\mformat&\underline{639\kformat}&\underline{2.17\mformat}\\{\bf \small{London}}
    &\verb//{\it BRISK}&1102&2.94\mformat&1.237\mformat&101.2\mformat&514\kformat&2.09\mformat\\(1,576)
    &\verb//{\it KAZE}&1068&2.75\mformat&1.237\mformat&110.9\mformat&617\kformat&2.15\mformat\\~\\
    &\verb//{\it TILDE}&697&1.85\mformat&1.234\mformat&81.6\mformat&323\kformat&2.01\mformat\\
    &\verb//{\it SuperPoint}&824&1.63\mformat&1.236\mformat&74.5\mformat&289\kformat&2.06\mformat\\
    &\verb//{\it D2Net}&924&2.37\mformat&1.237\mformat&\underline{114.2\mformat}&547\kformat&2.09\mformat\\~\\
    &\verb//{\it FFD}&\textbf{1151}&\textbf{3.56\mformat}&\textbf{1.239\mformat}&\textbf{117.3\mformat}&\textbf{688\kformat}&\textbf{2.23\mformat}\\
    \bottomrule
    \end{tabular}
    \end{minipage}
    \vspace{0.00mm}
\end{table*}
\enlargethispage{12pt}

We further evaluate the performance of the feature detectors for 3D reconstruction. According to the pipeline introduced in ~\cite{Schonberger}\footnote{\url{https://github.com/ahojnnes/local-feature-evaluation}}, the cameras are first calibrated in advance via Structure from Motion (SfM). Then, Multi-View Stereo (MVS) is applied to the output of SfM to obtain a dense reconstruction of the given scene. The quality of 3D models, which are the outputs of MVS, directly depends on the accurate and complete estimation of the camera parameters in the first step, i.e. SfM. We follow the same metrics and protocols~\cite{Schonberger} for analysing the 3D models. According to this paper, the SfM and MVS analyses are made via COLMAP~\cite{schonberger2016structure} and the metrics used are \textit{the number of registered images}, \textit{mean reprojection error}, \textit{the number of observations}, \textit{the number of inlier pairs and matches}, \textit{mean track length}, \textit{reconstructed sparse points} and \textit{reconstructed dense points}. The datasets employed here are Fountain, Herzjesu, Madrid Metropolis, Gendarmenmarkt and Tower of London. Exhaustive image matching was employed for all the datasets and they do not need image retrieval. 
Similar to the previous section, keypoints were detected by different feature detectors and then described by the HardNet++ descriptor. According to the pipeline~\cite{Schonberger}, the mutual nearest neighbours algorithm was employed for matching features.

\begin{figure*}
\begin{center}
\begin{minipage}[h]{1.0\linewidth}
    \centering
    \includegraphics[height=0.3\linewidth]{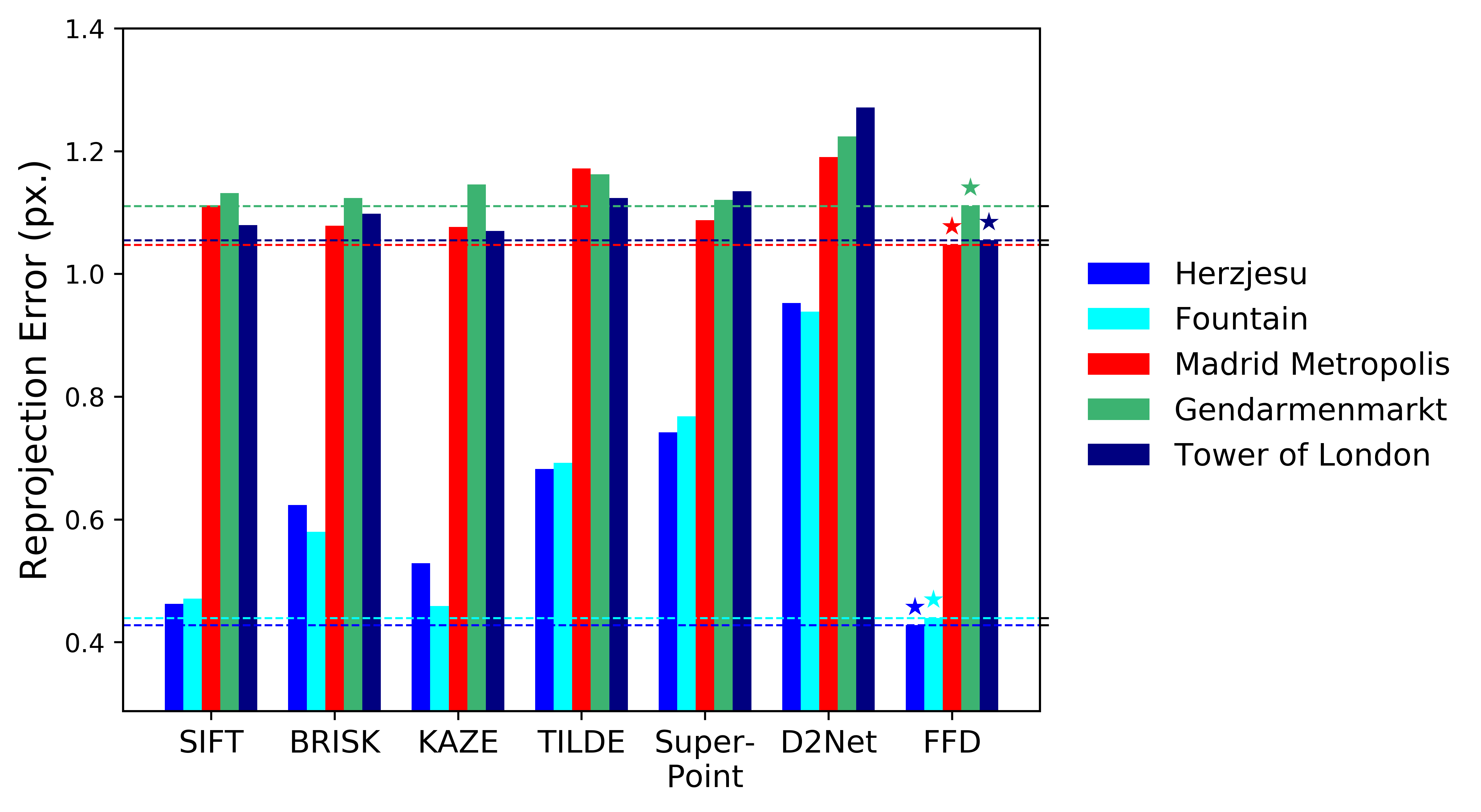}
    \includegraphics[height=0.3\linewidth]{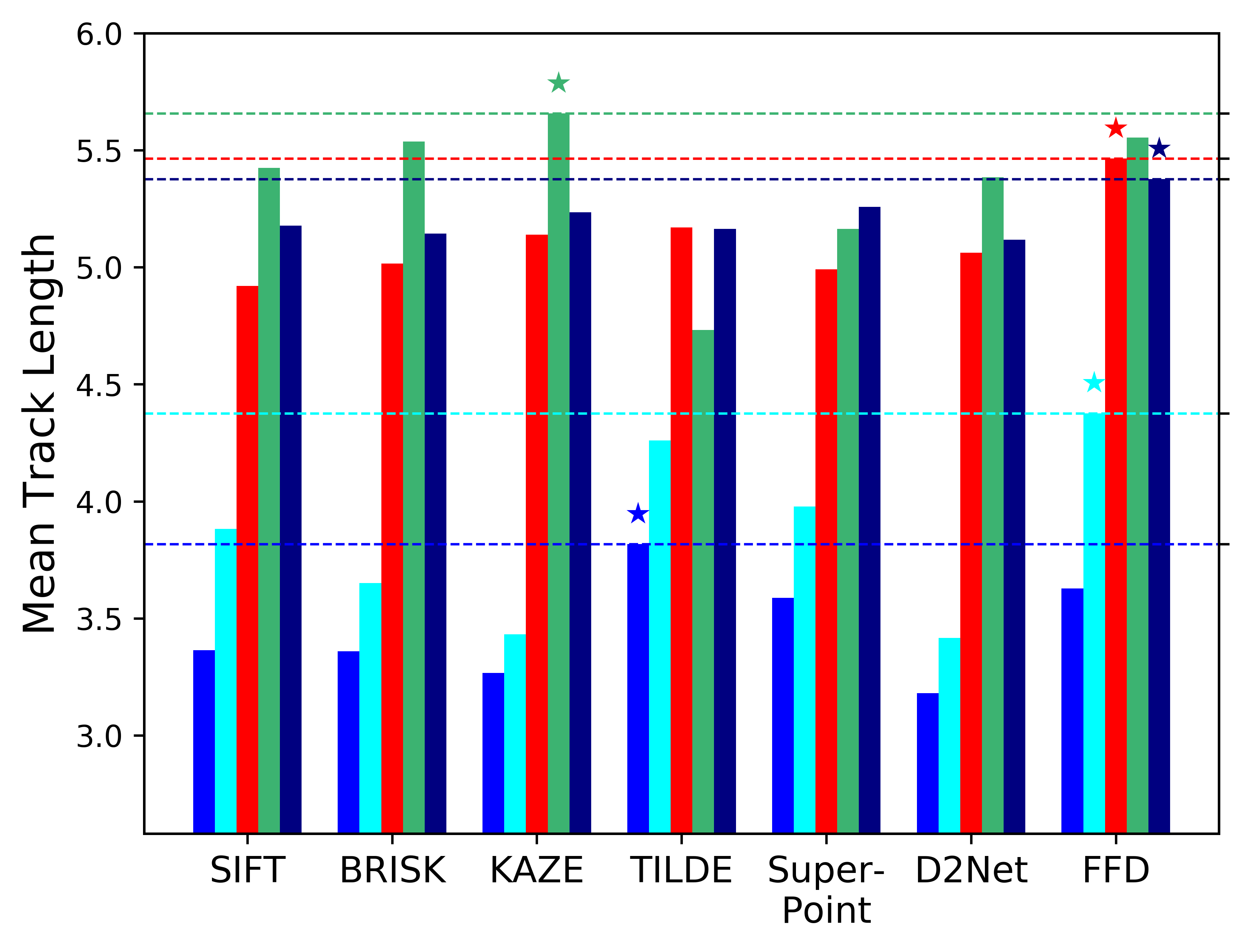}
    \caption{The reprojection error (left) and the mean track length (right) of different feature detectors for 3D reconstruction over different datasets. The best result in each dataset is highlighted with an asterisk.}
    \label{fig::3d_figs}
    \end{minipage}
    \vspace{0.00mm}
\end{center}
\end{figure*}

The quantitative results are reported in Table~\ref{table::results3dults} and Fig.~\ref{fig::3d_figs}. We reported the results of just six existing methods that gained the best performance in the previous sections.
For the two smaller datasets i.e. Fountain and Herzjesu, which are relatively easy benchmarks due to the structured camera setup with high overlap, FFD performs better than the existing feature detectors in terms of the number of observations, the number of inlier matches and the number of sparse points.
In the larger-scale datasets i.e. Madrid
Metropolis, Gendarmenmarkt and Tower of London, which are more challenging for 3D reconstruction due to large variations in illumination and viewpoint, FFD performs best among all the feature detectors, both in terms of sparse and dense reconstruction results. Our technique consistently produces the most complete sparse reconstruction results in terms of
the number of registered images and inlier pairs, resulting in the dense models including the most points because of accurate camera pose estimation.

According to Fig.~\ref{fig::3d_figs}, FFD generally performs on par with or better than the existing techniques in terms of mean track length.
The mean reprojection error shows that the multiscale techniques generally perform better than the learning ones. The localization errors of the proposed FFD are the lowest, indicating the highest precision of its detected keypoints.
These results are consistent with those reported in the previous section where FFD gained the highest performance in terms of localisation accuracy. 

\begin{figure}[h!]
\centering
    \includegraphics[width=0.78\linewidth]{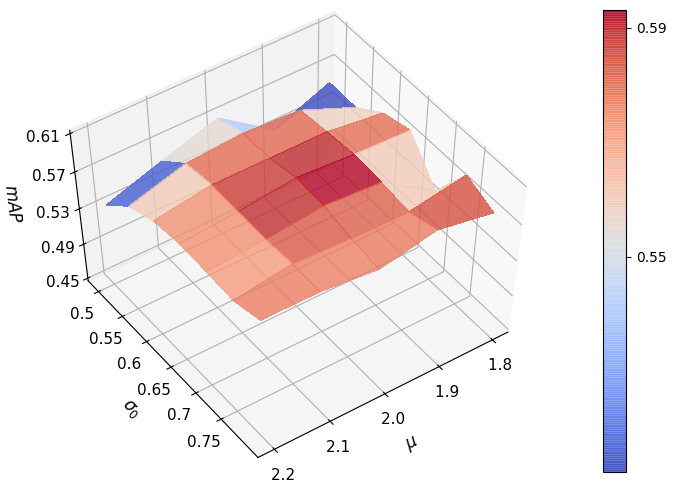}
    \includegraphics[width=0.8\linewidth]{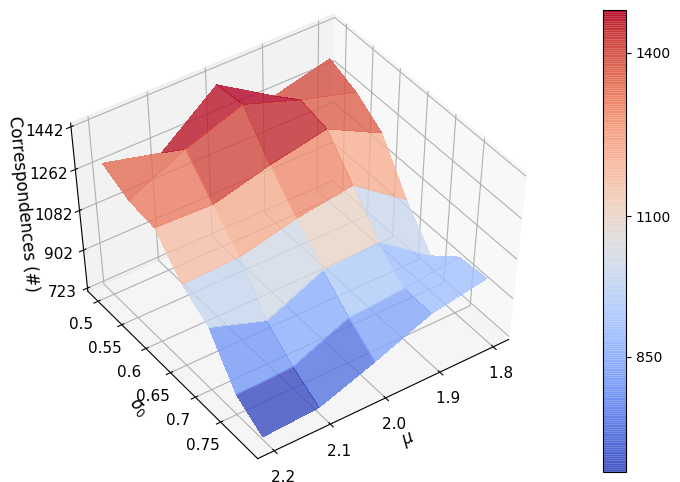}
    \caption{The influence of parameters $\sigma_0$ and $\mu$ inside the proposed FFD on the quality (mAP and \#correspondences) of the detected keypoints.}
    \label{fig:ablation}
\end{figure}

\begin{figure}
    \centering
    \subfigure[SIFT]{
    \includegraphics[height=1.5in]{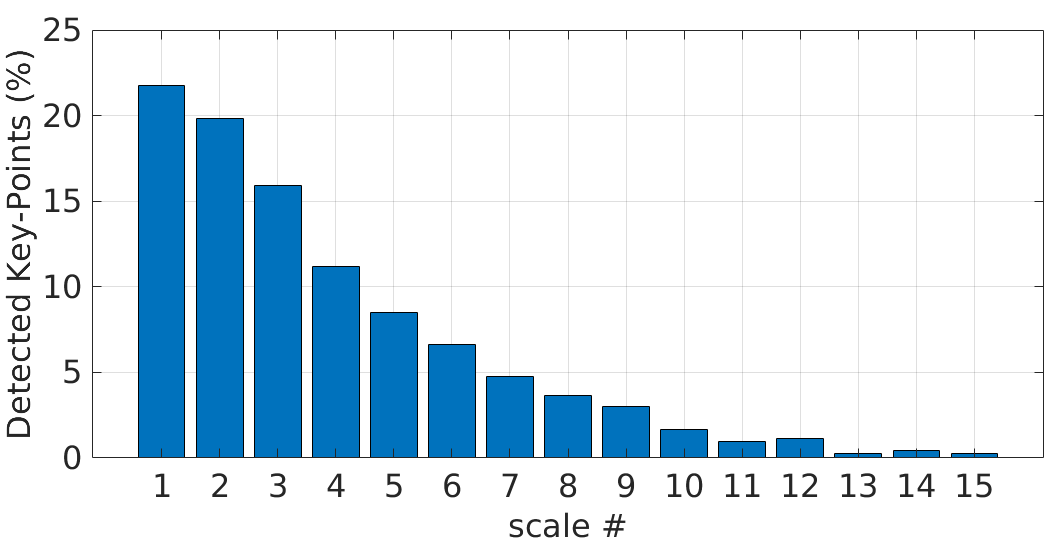}
    \label{fig::distributionsift}}
    \subfigure[FFD]{
    \includegraphics[height=1.5in]{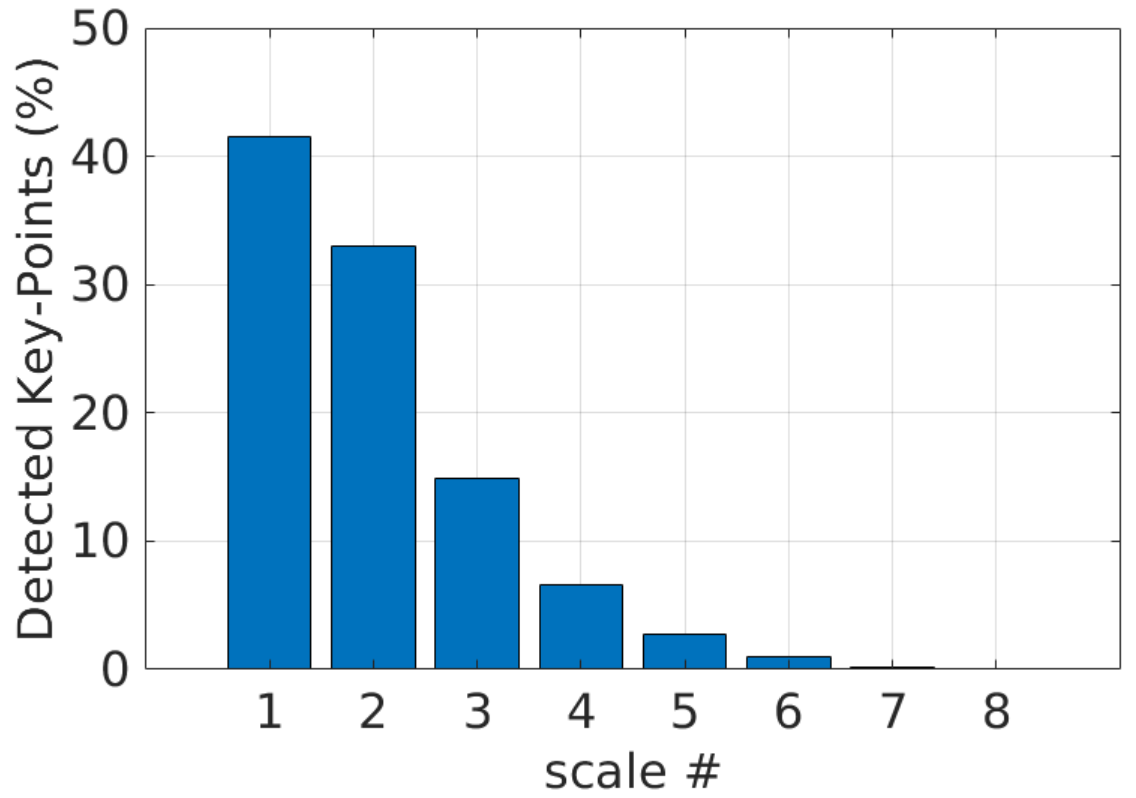}}
    \caption{The distribution of keypoints detected by SIFT and FFD across different scale levels.}
    \label{fig::distribution}
\end{figure}

\subsection{Golden parameter values}

 In this section, we carry out an ablation study about whether the golden values of the parameters $\sigma_0$ and $\mu$ in the proposed FFD are optimal. To this end, we reported its results about the mAP and the number of correspondences of the detected keypoints with different values of $\sigma_0$ and $\mu$ over the Illumination and Viewpoint datasets of the HSequence benchmark. The experimental results are presented in Fig.~\ref{fig:ablation}. The figure shows that for a fixed $\mu$, the mAP of the detected keypoints is improved by increasing $\sigma_0$ from 0.5 to 0.65 but further increasing it causes a serious decline in the number of the keypoints detected without enhancing their mAP.
Likewise, for a fixed $\sigma_0$, more deviation from $\mu=2$ results in a lower mAP and a smaller number of the correspondences. In short, if we take both these metrics into account, `$\mu$ around 2' and `$0.55\leq\sigma_0\leq0.65$' give stabler results.

\subsection{Distribution of Keypoints Across Scales}
Fig. \ref{fig::distribution} reports the distribution of the keypoints per scale to  the total number of those detected by SIFT and FFD. 
The distribution of the keypoints for SIFT [Fig.~\ref{fig::distributionsift}] is more even across the scales while FFD detects the majority of the keypoints at its first two scale levels (about three-quarters). There are less than one-sixth of the detected keypoints over the next scale and just less than 10\% of the whole keypoints are located over the other scales. This is because FFD has a larger sigma value that spans a larger area, leading the images to be smoothed more heavily and subject to more serious geometric distortion and even some artefacts around image edges; and thus fewer extreme blobs/patches in the fine scale-space pyramid to be identified. From the running time perspective, this trend is more favourable as `$N=3$' assures that the majority of reliable keypoints can be detected in the first two scale levels. Note that unlike the conventional feature detectors that split the input image into several octaves and each octave includes `$S$+3' scale levels, we decompose the given image into just `$N+3$' levels that includes all the octaves and scale levels.

\subsection{Run Time and Computational Complexity }
FFD was implemented in C++/OpenCV3.4 (without boost) and all the experiments were carried out on a 64-bits computer with Intel(R) Xeon(R) Gold 6130 CPU @ 2.10GHz processors, 48 GB RAM and two Tesla P100-PCIE-16GB GPU devices. 

The execution time of all the detectors as well as AKAZE is reported in Table \ref{table::runtime}. 
From the table, it can be seen that the computational time of HarrisZ and KAZE is high and the latter is markedly improved by its accelerated variant, i.e. AKAZE. Overall, SIFT and AKAZE need more running time while SURF and BRISK need almost one-third of that time. 
Although the computational time of the recent learning feature detectors like D2Net show promising results, their time-cost is still high and most of them require GPU platform. 
When we compare the running time of the fastest conventional feature detector, i.e. BRISK, our feature detector needs just about one-fifth of that computational time. BRISK uses the FAST feature detector at its heart that is a fast intensity-based detector. 
The computational time of FFD shows that it can solve the main drawback of the multiscale feature detectors, i.e. high computational cost. 
FFD reduces the running time of SIFT by about 95\%. It is worth reporting that in FFD, 54\% of the computational time is assigned to the construction of the scale-space pyramid, 25\% to non-maximum suppression, and the rest to the other steps including the refinement and edge suppression. These figures for SIFT are 74\%, 6\% and 20\%, respectively. As the majority of SIFT and FFD computational times is assigned to the pyramid construction and non-maximum suppression, we also analyze their theoretical computational complexities as follows. 

The complexity of the scale-space pyramid varies with the number of scales and we consider 2 comparable scale levels where SIFT gives its largest scale-ratio, or equivalently its fastest version. The number of octaves in SIFT is set to 4, where we have one upsampling and two downsampling operations. 4 octave levels in SIFT is equivalent to setting $N$ to 2 for FFD, as we have no upsampling operation. The length of the Gaussian filter $N_f$ is taken as `$7\sigma_i$'. Both of the feature detectors use separable convolution that needs $N_f$ multiplications and `$N_f-1$' additions. If we ignore the complexity of upsampling, SIFT needs `$955MN$' operations, where $M$ and $N$ are the dimensions of the input image. Unlike SIFT, the kernel set in FFD is fixed at length 5 and at each scale level, the stride is changed through inserting zero between its elements; in practice, changing stride does not need arithmetic operations but at the cost of the memory accesses only. The total number of the required operations by FFD to form its fine scale-space pyramid is `$49MN$', which is about 5\% of the operations required by SIFT. To analyze the complexity of the non-maximum suppression step, we disregard the pixels located at borders; this step takes `$10.625MN$' comparisons in SIFT and `$2MN$' for FFD. Bearing these matters in mind, FFD also detects highly reliable keypoints. These remarkable characteristics show that FFD is more suitable for real-time applications.

\begin{table}[!t]
\centering
    \caption{The computational time in milliseconds of different feature detectors over an image with $800\times640$ pixels (The average time over 20 runs is reported here).}
    \label{table::runtime}
    \begin{tabular}{lcccc} 
    \toprule
    {\bf Detector} & Category&Platform&{Run Time ($ms$)}\\ \midrule
    \verb//SIFT &Multiscale&CPU&552 \\ 
    \verb//SURF &Multiscale&CPU&159 \\ 
    \verb//BRISK &Multiscale&CPU&147 \\
    \verb//HarrisZ &Multiscale&CPU&2700\\
    \verb//KAZE  &Multiscale&CPU&1500 \\
    \verb//AKAZE &Multiscale&CPU& 438 \\~\\
    
    \verb//DNet&Deep learning&\textbf{GPU} &1300\\
    \verb//TILDE&Deep learning&CPU & 12100 \\
    \verb//TCDET&Deep learning&\textbf{GPU}&4100\\
    \verb//SuperPoint&Deep learning&\textbf{GPU}&54\\
    \verb//D2Net&Deep learning&\textbf{GPU}&950\\~\\
    \verb//FFD&Multiscale&CPU&29\\
\bottomrule
\end{tabular}
\end{table}
\enlargethispage{12pt} 

\section{Conclusion and Future Work}
\label{sec::conclusion}
In this study, we have proposed a novel detector, called fast feature detector (FFD). The main problems with conventional feature detectors are their scale-space analysis and computational burden. 
We have tackled these drawbacks by analysing the relations between LoG and DoG in scale normalization and excitatory regions, where the DoG is often used to approximate LoG for the sake of computational efficiency and the reduction of noise sensitivity. 
We proved that reliable scale-space pyramids in the continuous domain are obtained under a specific range of blurring ratios and smoothness widths that are presented in Fig.~\ref{fig::appendixBim}.
We also deduced that a blurring ratio of 2 and a smoothness width of 0.627 guarantee that the resulting pyramids enable adjacent edges in the given image 
to be as separable as possible. 
These golden values provide valuable knowledge and insights into the design of an appropriate kernel in the continuous domain, which is then discretized in order to make it applicable to discrete images using the undecimated wavelet transform and the cubic spline function. 
Experimental results and a comparative study with  state-of-the-art techniques over several publicly accessible datasets and example applications show that 
FFD can detect more highly reliable feature points in the shortest time, which makes it more suitable for real-time applications. 
Many real-time applications, like advanced driver assistance systems and 3D phenotyping of plants, require fast and robust feature detectors. 
Investigating the effectiveness of the proposed feature detector for such applications could be interesting.

\section*{Acknowledgments}
The authors gratefully acknowledge the HPC resources provided by Supercomputing Wales (SCW) and Aberystwyth University. MG acknowledges his DCDS and Presidents scholarships awarded by Aberystwyth University. YL is grateful to the partial funding by BBSRC and UKIERI through grants BB/R02118X/1 and DST UKIERI-2018-19-10 respectively. We thank the Associate Editor and three anonymous reviewers for their constructive comments that have improved the quality of the paper.

\vspace{-1.5cm}
\begin{IEEEbiography}
[{\includegraphics[width=1in,height=1.25in,clip,keepaspectratio]{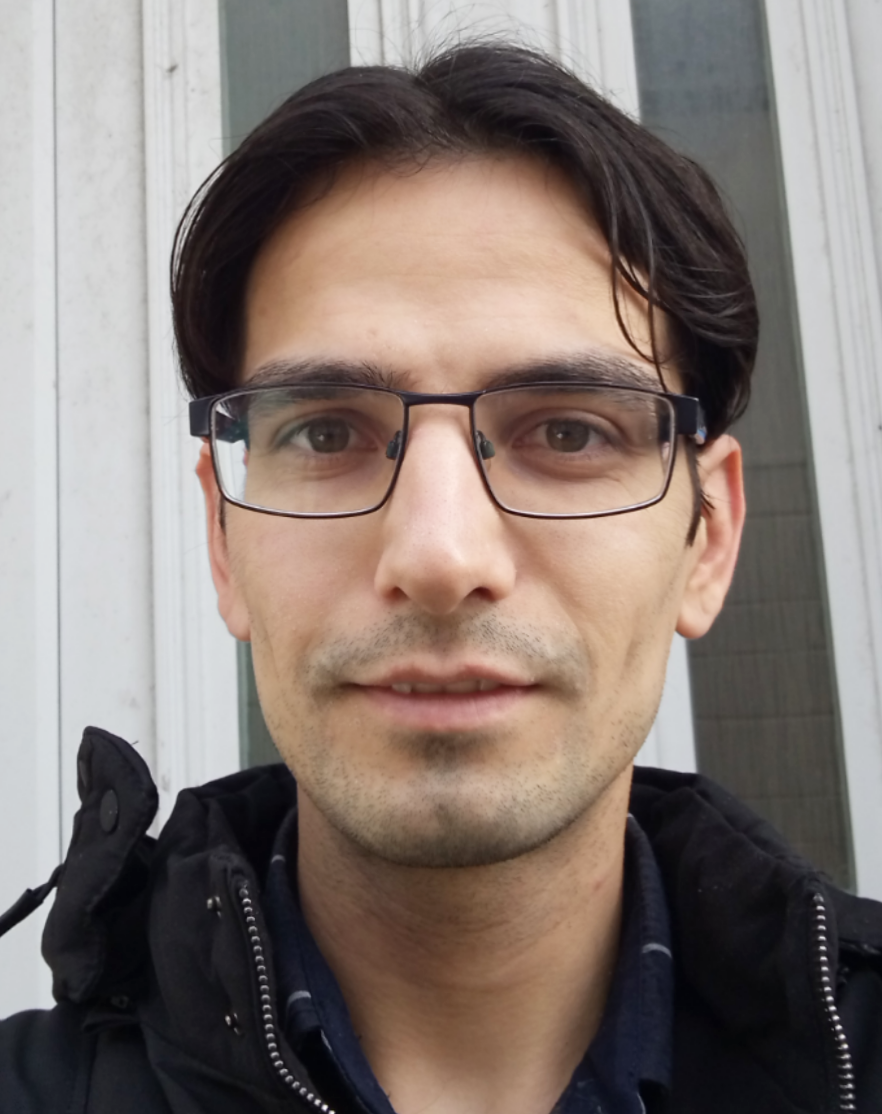}}]{Morteza Ghahremani} is working towards his PhD with a speciality in 3D computer vision at Aberystwyth University.
He is primarily focused on understanding and developing robust computational algorithms for analysing images. 
He has published several papers in the top-ranked international journals and conference proceedings and serves as a reviewer for IEEE Transactions on Pattern Analysis and Machine Intelligence, IEEE Transactions on Image Processing, IEEE/CVF Conference on Computer Vision and Pattern Recognition, etc. His research interests are structure from motion, 3D point cloud processing, image super-resolution and deep learning.
\end{IEEEbiography}
\vspace{-1.3cm}
\begin{IEEEbiography}[{\includegraphics[width=1in,height=1.25in,clip,keepaspectratio]{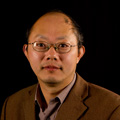}}]{Yonghuai Liu}
is a professor at Edge Hill University. He is an editor for several top-ranked international journals and conference proceedings in the field including Neurocomputing and Pattern Recognition Letters and International Conference on Robotics and Automation. He has published more than 190 papers in the top-ranked international journals and conference proceedings. His main research interests include 3D computer vision, image processing and machine learning.
\end{IEEEbiography}
\vspace{-1.3cm}
\begin{IEEEbiography}[{\includegraphics[width=1in,height=1.25in,clip,keepaspectratio]{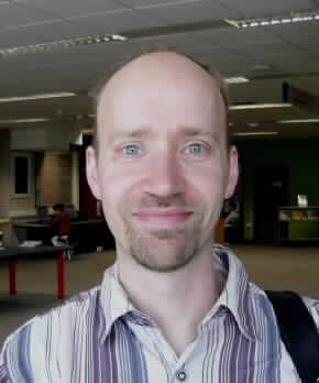}}]{Bernard Tiddeman}
is a Reader and former Head of Department in Computer Science at Aberystwyth University. He obtained his PhD from Heriot-Watt University in 1998. From 1999-2010 he worked as a researcher and then lecturer at the University of St Andrews. His research interests include: 2D and 3D facial image analysis and synthesis; computer vision for robotics; and vision and graphics applications in archaeological heritage.
\end{IEEEbiography}
\end{document}